\documentclass[journal]{IEEEtran} 
\IEEEoverridecommandlockouts                              % This command is only
\usepackage{graphicx}
\usepackage{amsmath}
\usepackage{cite}
\usepackage{url}

\usepackage{times}
\usepackage{fancyhdr,graphicx,amsmath,amssymb}
\include{pythonlisting}

\usepackage{algorithm}
\usepackage{algpseudocode}
\usepackage{soul}

\usepackage[font=footnotesize,belowskip=-0.5pt,aboveskip=0pt]{caption}
\usepackage{bigstrut}
\usepackage[utf8]{inputenc}
\usepackage{array}
\usepackage{multirow}
\usepackage{tabularx}
\usepackage{bigstrut}
\usepackage{makecell}
\usepackage{tabu}
\usepackage{bm}
\usepackage{caption}

\usepackage{amsmath}          
\usepackage{graphics}
\usepackage{graphicx}
\usepackage{float}
\usepackage{caption}
\usepackage{subcaption}
\usepackage{dblfloatfix}
\usepackage{calc}
\usepackage{url}
\usepackage{textcomp}
\usepackage{balance}
\usepackage{siunitx}
\usepackage{mdwlist}
\usepackage{comment}
\usepackage{enumerate}
\usepackage[shortlabels]{enumitem}

\usepackage{gensymb}

\usepackage{array}
\usepackage{multirow}
\usepackage{tabularx}
\usepackage{booktabs}

\usepackage{bigstrut}
\usepackage{fixltx2e}

\usepackage{color}
\usepackage{hhline}
\usepackage[dvipsnames]{xcolor}
\usepackage{soul}

\newcommand{\revision}[1]{\textcolor{black}{#1}}

\usepackage[capitalize]{cleveref}
\usepackage{fancyhdr}

\begin{document}

\title{\LARGE \bf
%We should have a new title I think, since authorship is different
%``The last centimeter:'' Adaptive gripping with a smart suction cup
Haptic search with the Smart Suction Cup on adversarial objects %improves grip success
%\\can improve pick-and-place

%A Multi-Chamber Smart Suction Cup \\for Adaptive Gripping and Haptic Exploration
%Suction cup tactile sensor for haptic exploration and localizing leaks
%Suction cup tactile sensor for haptic exploration and grip monitoring
%Suction cup tactile sensor for haptic exploration and dynamic contact monitoring
%For future versions: DexCup, etc. (or other name?) DexVac, HapVac 
}

\author{Jungpyo Lee$^{1*}$, Sebastian D. Lee$^{1*}$, Tae Myung Huh$^{2}$, Hannah S. Stuart$^{1}$% <-this % stops a space
\thanks{\textbf{Accepted final version.} To appear in IEEE Transaction on Robotics, 2023. © 2023 IEEE.  Personal use of this material is permitted.  Permission from IEEE must be obtained for all other uses, in any current or future media, including reprinting/republishing this material for advertising or promotional purposes, creating new collective works, for resale or redistribution to servers or lists, or reuse of any copyrighted component of this work in other works.
}
\thanks{This paper has supplementary downloadable material available at http://ieeexplore.ieee.org, provided by the authors. This includes video clips of haptic exploration.}%
\thanks{$^{1}$J. Lee, S. D. Lee and H. S. Stuart are with the Embodied Dexterity Group, Dept. of Mechanical Engineering, University of California Berkeley, Berkeley, CA, USA. {\tt\small jungpyolee@berkeley.edu; sebastiandavidlee@berkeley.edu; hstuart@berkeley.edu}}
\thanks{$^{2}$T. M. Huh is with the Dept. of Computer and Electrical Engineering, University of California Santa Cruz, Santa Cruz, CA, USA. {\tt\small thuh@ucsc.edu}}%
\thanks{$^{*}$J. Lee and S. Lee contributed equally to this work.}
}

% The paper headers
% The paper headers
% \markboth{IEEE Transactions on Robotics,~Vol.~X, No.~X, Month~2023}%
% {Huh \MakeLowercase{\textit{et al.}}: A multi Chamber Smart Suction Cup}

\maketitle

% \thispagestyle{empty}
% \pagestyle{empty}

% \global\csname @topnum\endcsname 0
% \global\csname @botnum\endcsname 0

%%%%%%%%%%%%%%%%%%%%%%%%%%%%%%%%%%%%%%%%%%%%%%%%%%%%%%%%%%%%%%%%%%%%%%%%%%%%%%%%
\begin{abstract}

%it says the limit is 200 words here: https://www.ieee-ras.org/publications/t-ro/information-for-authors
%this abstract is exactly 200 words right now
%Suction cups are an important gripper type in industrial and logistics robot applications, and prior literature focuses on using vision-based planners to improve grasping success in these tasks. However vision-based planners can fail due to adversarial objects or lose generalizability when applied to an unseen scenario without retraining learned algorithms. We propose haptic exploration to improve suction cup grasping when visual grasp planners fail. We present the Smart Suction Cup, an end-effector that utilizes internal flow measurements for tactile sensing. We show that model-based haptic search methods, guided by these flow measurements, improve grasping success by up to 2.5x as compared with using only a vision planner during a bin-picking task. In characterizing the Smart Suction Cup on both geometric edges and curves, we find that flow rate can accurately predict the ideal motion direction even with large postural errors. Critically, the Smart Suction Cup includes no electronics on the cup itself, such that the design is easy to fabricate and haptic exploration does not risk damaging the sensor. By focusing on adversarial scenarios, that might currently limit the use of vision-based planners, this work motivates the use of suction cups with autonomous haptic search capabilities to bridge adoption gaps.
Suction cups are an important gripper type in industrial robot applications, and prior literature focuses on using vision-based planners to improve grasping success in these tasks. Vision-based planners can fail due to adversarial objects or lose generalizability for unseen scenarios, without retraining learned algorithms. We propose haptic exploration to improve suction cup grasping when visual grasp planners fail. We present the Smart Suction Cup, an end-effector that utilizes internal flow measurements for tactile sensing. We show that model-based haptic search methods, guided by these flow measurements, improve grasping success by up to 2.5x as compared with using only a vision planner during a bin-picking task. In characterizing the Smart Suction Cup on both geometric edges and curves, we find that flow rate can accurately predict the ideal motion direction even with large postural errors. The Smart Suction Cup includes no electronics on the cup itself, such that the design is easy to fabricate and haptic exploration does not damage the sensor. This work motivates the use of suction cups with autonomous haptic search capabilities in especially adversarial scenarios.

\end{abstract}

%Official key words, ranked in suggested order:
%Best fit --> to input while submitting paper to inform reviewer selection:
%Grippers and Other End-Effectors --> Manipulation and Grasping 
%Force and Tactile Sensing --> Manufacturing, Process, and Service Automation
%Soft Robot Applications --> soft robotics (I think we should get reviewed my so soft mechanisms/design folks)
%Sensor-based Control --> Vision and Sensor-Based Control

%Secondary fit:
%Compliant Joints and Mechanisms --> Mechanisms, Design, and Control 
%Embedded Systems for Robotic and Automation --> Robotic systems 
%Perception for Grasping and Manipulation --> Autonomy for Mobility and Manipulation
%Product Design, Development and Prototyping --> Applications
%Failure Detection and Recovery --> Mechanisms, Design, and Control
%Soft Sensors and Actuators --> soft robotics
%Soft Robot Applications --> soft robotics

%%%%%%%%%%%%%%%%%%%%%%%%%%%%%%%%%%%%%%%%%%%%%%%%%%%%%%%%%%%%%%%%%%%%%%%%%%%%%%%%

% You can make your own Tex to make your paragraph or sections.

\section{Introduction}

%%%%%
\begin{figure}[tbp!]
\centering
	% \vspace{-10pt}
	%\begin{subfigure}[h]{0.5\textwidth}
	%\centering
	\includegraphics[width=0.9\linewidth]{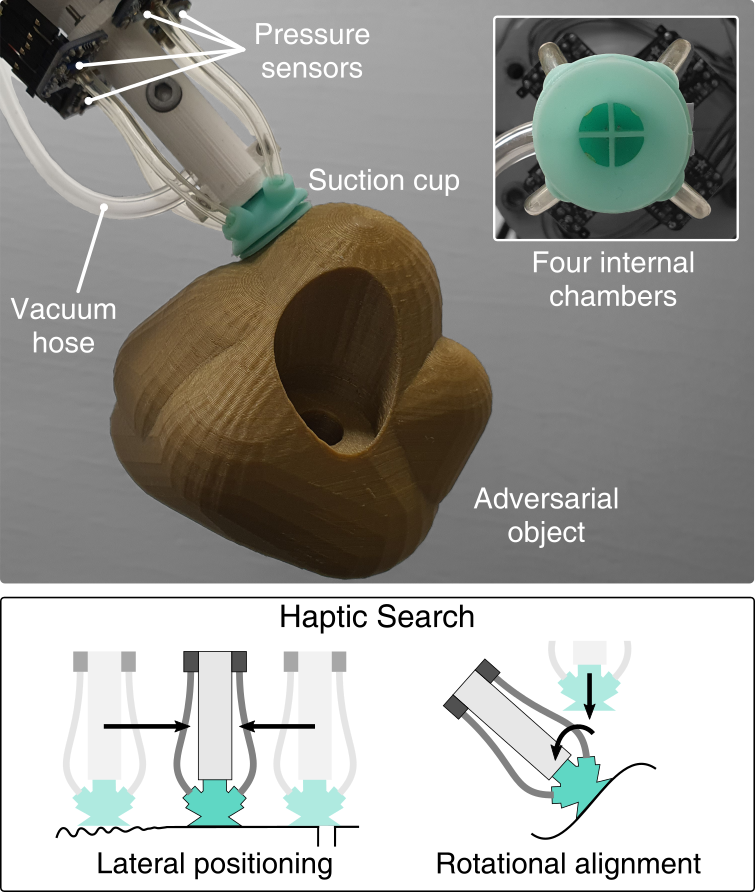}
% 	\includegraphics[width=1\linewidth]{./Figures/1_frontImage.png}
	%\end{subfigure}
    \vspace{10pt}
	\caption{The multi-chamber Smart Suction Cup grips an adversarial object. The cup has four internal chambers, each connected to a pressure transducer that provides a measure of internal flow rate. It is able to localize small breaks in the seal due to, for example, the rugosity (e.g., wrinkles, bumps, etc.) of the object surface. Haptic search can allow for successful gripping even when the initial grasping point fails, important for visually-adversarial objects. }
	\label{fig:main}
	% \vspace{-15pt}
\end{figure}
%%%%%%

Vacuum grippers, or suction grippers, are widely used in industry for simple pick and place operations. Relying on negative internal pressure that forms when sealed against a surface, the suction gripper can gently handle an object without applying squeezing force, which allows an astrictive handling of various types of objects. If the item to be grasped is smooth and well modelled, as in manufacturing lines, the gripper can repeatably and predictably handle it with high reliability. However, for grasping in unstructured environments, e.g., in e-commerce warehouses, objects vary dramatically and present many different surface conditions that may or may not be easy to visually perceive or grip with a suction cup. Careful planning of grasp contact location is therefore important, and methods for doing so have been widely studied for the past few years. 
While there have been successful demonstrations of versatile suction grasp planners, these methods often rely on vision, which may not capture fine object details of the geometry and lead to suction failure. Moreover, pre-trained models are typically specific to certain suction cup and camera configurations, making it challenging to transfer these methods to different hardware setups without retraining. Time-consuming retraining currently presents a barrier to adoption.

To address these challenges, we propose the use of autonomous haptic search -- or the repositioning of the cup using contact measurements -- to supplement vision in suction grasping. This new approach leverages pre-trained vision-based grasp planners to obtain an approximate solution before then %and incorporates robotic haptic exploration to 
fine-tuning the pose after contact occurs until the grasp succeeds. %contacts around the planned location and ensure suction grasp success. 
For this method to be effective, we assume that a successful grasp point %e believe that the optimal suction contact point 
is close to the pre-trained planner's solution even when errors emerge, as the planner already considers key factors of graspability such as the object's weight distribution and the suction seal formation of a similar suction cup. To adjust the contact location, we use haptic exploration driven by flow-based tactile sensors on our Smart Suction Cup, first presented in \cite{huh2021}. This design has the advantage of no electronics embedded in the cup itself, but remote sensors can still provide valuable information about local suction leakages to overcome grasp failures. %counteract the detected leakage to achieve a vacuum seal.

\subsection{Overview}

Section \ref{sec:relatedWorks} provides a review of related works.
In Section \ref{sec:Design}, the Smart Suction Cup is described along with computational fluid dynamics models to demonstrate the expected signals; this design and flow analysis was previously presented in our prior work \cite{huh2021}. %, and overlaps substantially with Section \ref{sec:Design}. 
In the current work, we evolve this concept substantially beyond the prior work by now introducing and implementing autonomous haptic search. %performing new complete characterizations of lateral positioning and rotational alignment procedures to guide autonomous haptic search. 
%While \cite{huh2021} focused on the prediction of grasp failures after the seal was formed on a smooth flat plate, the present work now focuses on first achieving a suction seal on more adversarial objects.
Section \ref{sec:algorithms} presents our new proposed haptic search algorithm that utilizes the flow readings to improve grasping on adversarial objects. %We also present the experimental control conditions. 
Experimental setup and procedures are described in Section \ref{sec:methods}, including both sensor characterization on \revision{primitive fixed} objects and %loose objects in 
a bin-picking task with loose \revision{adversarial} objects. Section \ref{sec:Result} presents the results of these experiments; overall, we find that the use of the Smart Suction Cup haptic algorithm provides useful controller estimates and more successful grasping. 
Discussed in Section \ref{sec:discussion}, the model-based haptic exploration encounters failure modes that can be further improved in future work.

The contributions of this paper are as follows:
\begin{enumerate}
  \item Presentation and characterization of the first Smart Suction Cup that can sense local suction seal leakage on flat and curved surfaces by using remote pressure sensors.
  \item Design of a suitable model-based haptic search controller using tactile sensing feedback to improve suction seal in real time.
  \item Bin-picking experiments to evaluate performance across adaptive control algorithms with comparison to an existing vision-based grasp planner.
\end{enumerate}

\section{Related works}
\label{sec:relatedWorks}

\subsection{Suction grasp planning using vision}

One major challenge in suction grasping is how to plan a contact location. Examples of planning methods include the heuristic search for a surface normal\cite{morrison2018cartman} and neural network training of grasp affordance using binary success labels\cite{zeng2022robotic}. Wan et al. (2020) use CAD model meshes to plan a grasp resisting gravitational wrench\cite{wan2020planning}, and Dex-Net 3.0 learns the best suction contact pose from a point cloud considering both suction seal formation and gravitational wrench resistance\cite{mahler2018dex}. Using a similar approach to Dex-Net, Cao et al. (2021) built a larger suction grasp dataset including RGB images and annotations of a billion suction points\cite{cao2021suctionnet}. Using physics simulation, Shao et al. (2019) demonstrated a self-supervised learning method that finds suction grasp policies from RGB-D images for cluttered objects\cite{shao2019suction}, and Cao et al. (2022) improved it by implementing dense object descriptors\cite{cao2022reinforcement}. These aforementioned methods rely on RGB or depth sensors, which may not perceive fine details critical to suction success, e.g., texture, rugosity, porosity, etc. Vision can also become occluded in cluttered environments and heavily distorted with reflective or transparent objects. % This leads to an tactile sensation that explores the best.
% \begin{itemize}

%   \item review some of the work done by Alberto Rogriguez, Ken Golberg (Dex-Net), and other groups. Some can be found in the related work section of the Dex-net 3.0 paper and the recent suction-million paper.
  
%   \item Our claim here is that we vision is good, but has limitations in its resolution and can be easily distorted (e.g., reflective material). Moreover, when the end-effector reaches the object, then the object is highly occluded so we can not rely on vision when we have to adjust the suction cup grasping.
% \end{itemize}

\subsection{Suction cup tactile sensors}

Prior tactile sensors designed for use in suction cups provide partial information about object properties and vacuum sealing state. Researchers employ strain sensors on a suction cup by coating PEDOT \cite{aoyagi2020bellows} or carbon nanotube \cite{lee2021electronically}, or by installing microfluidic channels filled with carbon grease \cite{shahabi2023octopus}. These strain sensors measure suction deformation during surface contact, estimating the compression forces and load distributions of suction cups \cite{aoyagi2020bellows}, surface angles and stiffness \cite{shahabi2023octopus}, and object weight and center of gravity \cite{lee2021electronically}. Alternatively, the contact of the suction cup can be measured indirectly by proximity sensors, including a capacitive base plate \cite{doi2020novel}, inserted fiber optic cable \cite{sareh2017anchoring}, and micro-LIDAR \cite{frey2022octopus}. However, these methods provide information about the cup deformation and surface proximity, which may not always correspond to a suction seal formation that is subject to fine local geometry and porosity. For direct contact sensing, Muller et al. (2017) report a thin pressure sensor array attached to the suction cup lips, measuring the distributed contact pressures \cite{muller2017sensor}. However, the sensor film on the contact layer may weaken the suction seals. 

Another straightforward approach is to monitor the internal vacuum pressure of the suction cup as a discrete measure of suction sealing, as in \cite{eppner2016lessons}. However, this prior implementation method does not localize the source of a leak around the lip's edge or measure local surface geometry, which is critical for adaptive haptic exploration for a better grasp. 

% on the suction cup to measure the deformation during surface contacts. The outer surface of a suction cup can be coated by PEDOT\cite{aoyagi2020bellows} or carbon nanotube to measure the  
% Aoyagi et al. coat piezoresistive polymer on a bellows suction cup to measure compression forces \cite{aoyagi2020bellows}. Lee et al. also coat the outer surface of a suction cup inspired by octopus, able to estimate the object's weight and center of gravity. 

% Doi et al. implement a capacitive proximity sensor on the base plate of a suction cup end-effector to measure the distance from the plate to the object surface\cite{doi2020novel}. These methods measure vacuum state indirectly from the deformation of the suction cup and proximity to the object. 

% Another straightforward approach is to monitor internal vacuum pressure of the suction cup as a discrete measure of suction sealing, as in \cite{eppner2016lessons}. None of these methods localize the source of a leak or measure local surface geometry. Nadeau et al. demonstrate how continuous suction-flow-rate measurement at the fingertips of a multifinger robotic hand can inform grasping and in-hand manipulation under water and note sensitivity to contact geometry \cite{nadeau2020tactile}, but do not address application to suction cups. 

\begin{figure*}[b!]
\centering
	%\vspace{-10pt}
	%\begin{subfigure}[h]{0.5\textwidth}
	%\centering
	\includegraphics[width=1\linewidth]{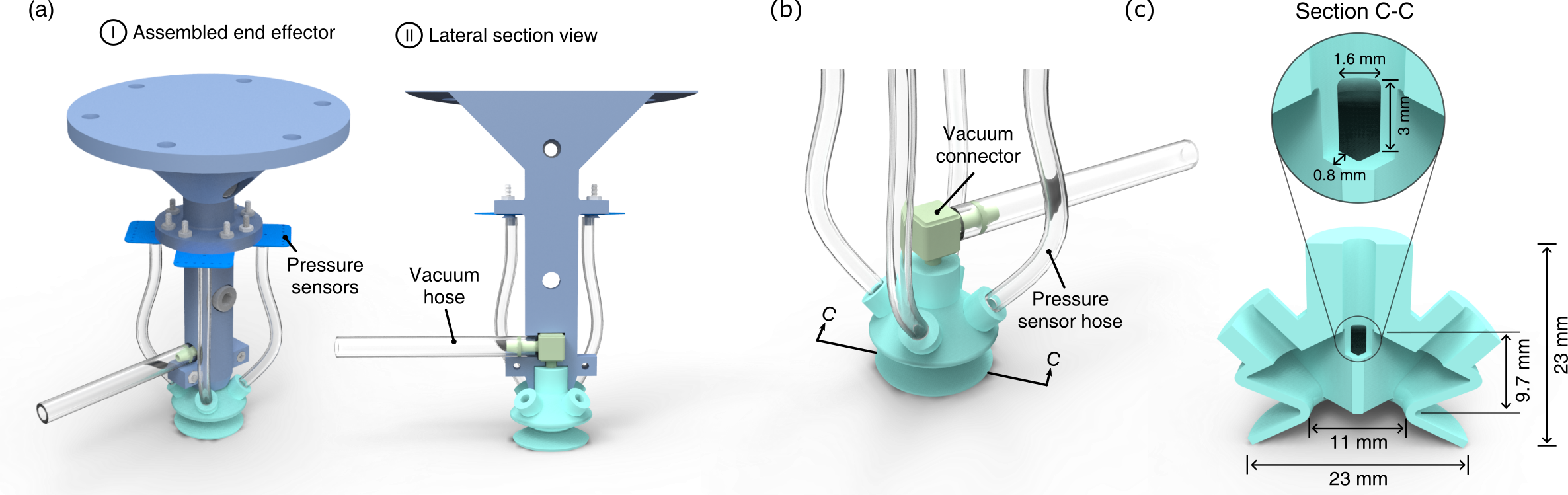}
	%\end{subfigure}
    \vspace{-5pt}
	\caption{Design of the end effector and the suction cup. (a) The end effector integration with the suction cup. (b) A close up of the suction cup shows how it is connected with a vacuum connector and hoses to the pressure sensors. (c) Cross-sectional view of the suction cup shows internal and outer dimensions.}
	\label{fig:cupDetail}
	 
%\end{figure*}
\vspace{15pt}
%\begin{figure*}[t!]
\centering
	%\vspace{-10pt}
	%\begin{subfigure}[h]{0.5\textwidth}
	%\centering
	\includegraphics[width=1\linewidth]{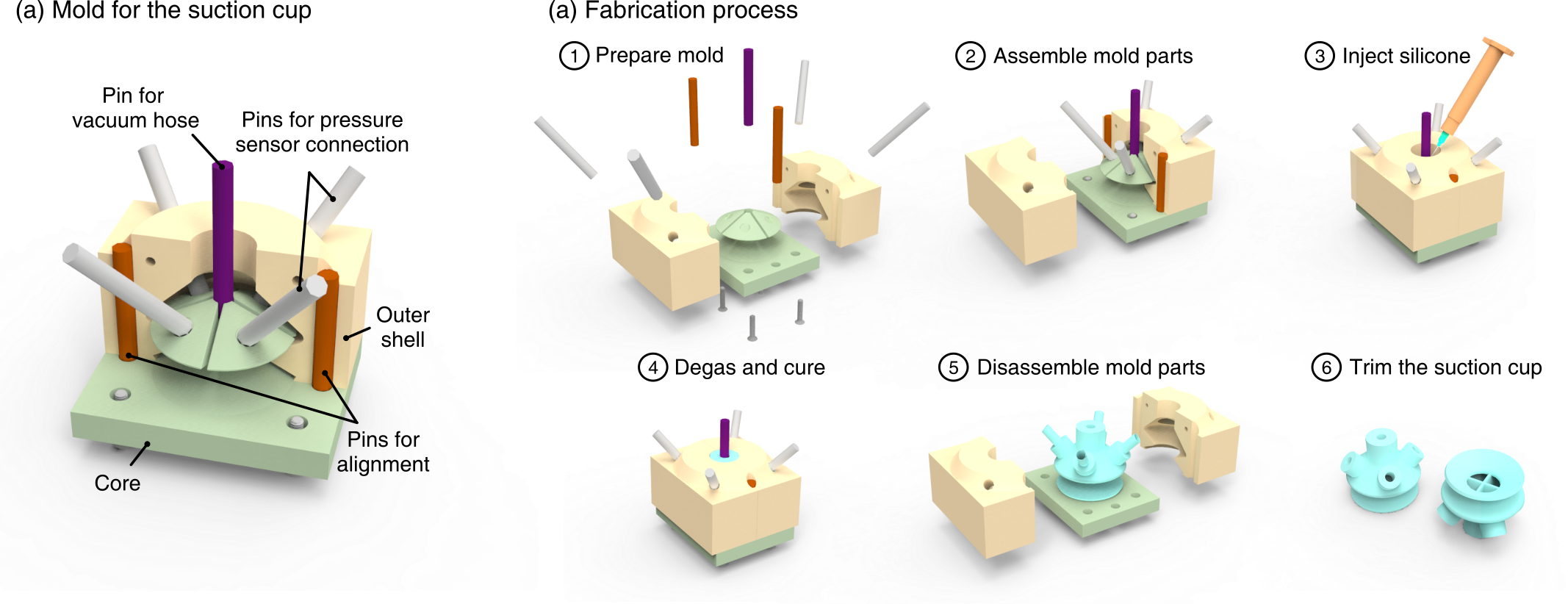}
	%\end{subfigure}
    \vspace{-10pt}
	\caption{Casting mold and fabrication of the suction cup. (a) The casting mold has three parts (2 Outer shells and 1 core). Molds are aligned and fixed by pins and bottom bolts. (b) The fabrication process of the suction cup.}
	\label{fig:cupFabrication}
	% \vspace{-15pt}
\end{figure*}

% Robotic Haptic Exploration
\subsection{Adaptive Regrasping using Tactile Sensing}

Robust grasping in real-world scenarios has driven research in adaptive regrasping using tactile sensing. Due to uncertainties in vision systems and difficulties capturing detailed object features, tactile sensors are employed to detect contact information and guide improvements in response to unsuccessful grasps. 
Adaptive regrasp research has predominantly focused on friction-based grippers rather than suction grippers. Simple regrasping approaches include increasing grasp forces or grasp impedance upon detection of perturbations, such as external forces causing slips\cite{romano2011human, hang2016hierarchical}. For multi-finger grippers, researchers have demonstrated finding better grasping points through finger gaiting \cite{hang2016hierarchical}. These methods primarily aim to improve handling or increase the stability of objects already held by the gripper.
In object-picking processes, deep learning or reinforcement learning techniques have been employed to process complex tactile sensor data. Chebotar et al. (2016) used a multi-finger gripper with a BioTac sensor to demonstrate regrasping of a simple cylindrical object; they analyzed complex spatiotemporal tactile sensor information with PCA and learned a regrasp policy to update the pose\cite{chebotar2016self}. Reinforcement learning was also used to learn hand grasping and regrasping policies in simulation, which are then effectively transferred to real robots \cite{wu2019mat}.
For parallel jaw grippers, vision-based tactile sensors, such as Gelsight, have been used\cite{hogan2018tactile,calandra2018more}. In \cite{hogan2018tactile}, the researchers trained a grasp quality metric from a given tactile image and simulated possible image shifts to guide the best regrasping policy. In \cite{calandra2018more}, they directly trained for the best action to achieve the highest grasp success, which could be either a regrasp or pick.

The majority of the approaches mentioned above rely on tactile sensing information processed by deep learning or reinforcement learning algorithms. These methods can be unintuitive and may require significant training data for generalization. These approaches may involve fully reopening the gripper during regrasp actions, which can be time-consuming. Moreover, theses approaches may not be applicable to suction grasping due to differences in grasping mechanisms.
In the following sections, we will present a physics- or intuition-based regrasping controller for suction cup grippers, enabling generalization without requiring extensive training data. Our controller operates without losing contact, potentially reducing operation times. To our knowledge, no existing literature addresses adaptive regrasping for suction cup grippers.

\begin{figure*}[h]
\centering
	%\vspace{-10pt}
	%\begin{subfigure}[h]{0.5\textwidth}
	%\centering
	\includegraphics[width=0.9\linewidth]{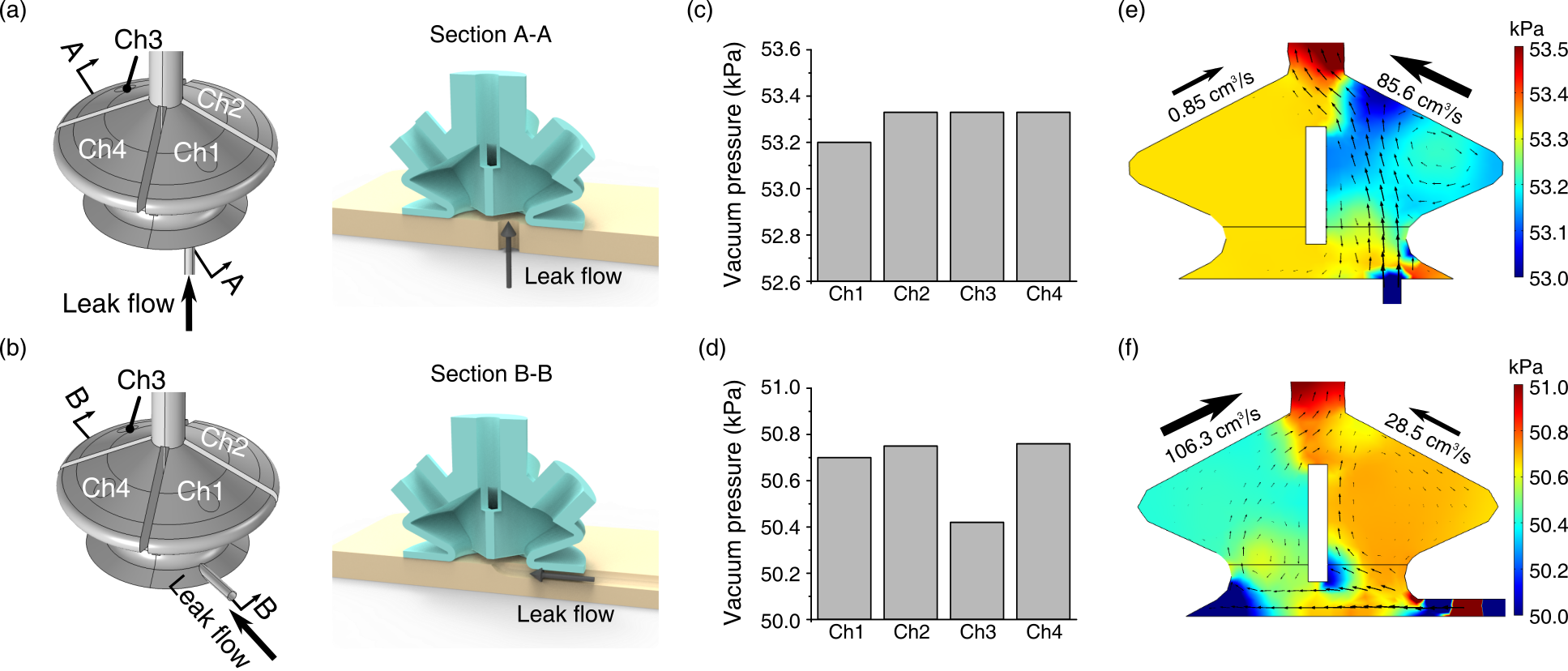}
	%\end{subfigure}
    \vspace{+3pt}
	\caption{(a-b) Two cases of CFD simulation. Light yellow blocks are engaged objects and the cross-sectional view shows leak flow into channel number 1. (c-d) CFD result of the vacuum pressure measured at the sensor locations of each chamber. The bar graphs are from the maximum of the four vacuum pressures. (e-f) Cross-sectional view of the pressure distribution. The arrows inside represent the relative logarithmic scale of airflow velocity.}
	\label{fig:cupCFD_simulation}
	% \vspace{-15pt}
\end{figure*}

\section{The Smart Suction Cup}
\label{sec:Design}

The Smart Suction Cup utilizes internal airflow estimates to monitor local contact conditions. %Hannah: let's consolidate this citation in the introduction. %Suction water flow rate could estimate contact states (e.g., the contact angle of a fingertip) of an underwater suction gripper \cite{nadeau2020tactile}, and we extend this concept to the air suction flow. 
Internal wall structures separate the internal cavity of the suction cup into four chambers (\cref{fig:main}) -- one for each cardinal direction. Overall suction airflow is therefore separated between each chamber and the pressure sensor connected to each chamber provides an estimate of the local flow rate. %Studying how spatial resolution and gripping performance is affected by altering the number of chambers will be a part of future work. 
We implement the wall structure inside a single-bellows suction cup %(\cref{fig:cupFabrication}) 
for its versatility on different curvatures and orientations of objects. The internal wall structure only spans the proximal portion of the suction cup, in order to maintain typical flexibility, deformation and seal formation at the distal lip. 
As shown in \cref{fig:cupDetail}a-b, the suction cup is mounted to an end effector fixture piece and connected with pressure transducers and a single vacuum hose with pressure regulation. For experimental trials, this end effector is integrated with a universal robot arm. 
Dimensions and internal geometry of the compliant cup are shown in \cref{fig:cupDetail}c.
A single prototype is used throughout experimental testing, without incurring damage or needing replacement.

\subsection{Fabrication}

We fabricate this 3D rubber structure including the chamber walls as in \cref{fig:cupFabrication}, with a single-step casting of silicone rubber. The casting mold comprises three parts, two outer shells and one core, that are 3D printed using an SLA 3D printer (Formlabs, Form2). These are assembled together using stainless steel dowel pins and bolts. To ensure the clean casting of the thin internal wall structures (0.8 mm thick), we used a syringe with a blunt needle (gauge 14) to inject uncured RTV silicone rubber (Smooth-On, MoldMax 40) and then vacuum-degassed it. After curing, the outer shells are removed and the silicone suction cup is stretched and peeled off of the inner core mold. Tearing of the silicone can occur during this step, especially with harder rubbers. Cast flashing around the lip of the cup can occur at the interface between the core and outer shells; deflashing is performed manually after demolding using a razor blade.

\subsection{CFD Simulation}
% \label{sec:DesignCFD}

Using Computational Fluid Dynamics (CFD) simulation (COMSOL Multiphysics, $k-\epsilon$ turbulence model), we evaluate the gripper in two example suction flow cases: \textit{vertical} and \textit{horizontal} flow (\cref{fig:cupCFD_simulation}a and b, respectively). The vertical flow case emulates when the suction cup only partially contacts a surface, or when the surface's shape inhibits sealing. %In haptic explorations, especially during geometric search, we surmise that the contact will often be partially made, flowing the leakage air vertically. 
However, when the suction cup engages with a smooth flat surface, flow can only move inward from the outer edges of the cup, as in the horizontal flow case. This horizontal leak is common as the suction cup is wrenched from the surface after a suction seal is formed. %, the vacuum seal is assumed to break from a single direction, flowing the leakage air horizontally along the contact surface. 
Although the suction cup will deform under vacuum pressure, we use modeled rigid geometry in the CFD simulation. For each case, we %used the same geometry as our designed suction cup and 
approximate the leak flow direction with a small pipe (D = 1 mm, L = 7 mm) intersecting with one of the internal chamber volumes as shown in \cref{fig:cupCFD_simulation}a-b. The boundary conditions of the vacuum pump pressures and flow rates match the experimental setup.

The simulation results suggest that the gripper can detect leakage flows using differences between the four pressure transducers. We defined vacuum pressure ($P_{vac}$) as 
\begin{equation}
% \vspace{-15pt}
P_{vac} = P_{atm} - P_{chamber}
% \vspace{+5pt}
\label{eqn:P_vac_Def}
\end{equation} \revision{where $P_{atm}$ is atmospheric pressure.}
In the vertical leakage flow case, $P_{vac}$ close to the leaking orifice shows the least vacuum pressure than the others (\cref{fig:cupCFD_simulation}c). On the other hand, the horizontal leakage causes the diagonally opposite channel %across the center 
to have the lowest $P_{vac}$ (\cref{fig:cupCFD_simulation}d). %These two different pressure distributions are caused by the flow rates through each chamber. 
These trends are supported by the flow results in \cref{fig:cupCFD_simulation}e-f, where the vertical and horizontal orifices produce the highest flow rate in opposite chambers.
%In the vertical leakage case, as shown in \cref{fig:Vertical2D}, the flow mostly passes through the orifice chamber while in the horizontal case (\cref{fig:horizon2D}), more leakage flow passes through the chamber across due to the flow path along the bottom surface. 
The simulation result also shows an estimate of the pressure difference between chambers ($\sim$0.4kPa) which must be differentiated by the selected pressure sensors.
% \revision{Despite the two distinctive results of these idealized simulations, %the actual pressure distributions may not be interpretable due to the 
% real suction cup deformation and complex leak geometry necessitate data-driven analysis of pressure as a tactile sensor as described in \cref{sec:Detach}.}

\subsection{System integration}
% \label{sec:system}

%\han{A little confusing, so I tried rewording based on my understanding. Please correct new mistakes. }

%We integrated the fabricated suction cup with a vacuum generator, pressure sensors, and a robot arm. 

\begin{figure}[t]
\centering
	%\vspace{-10pt}
	%\begin{subfigure}[h]{0.5\textwidth}
	%\centering
	\includegraphics[width=1\linewidth]{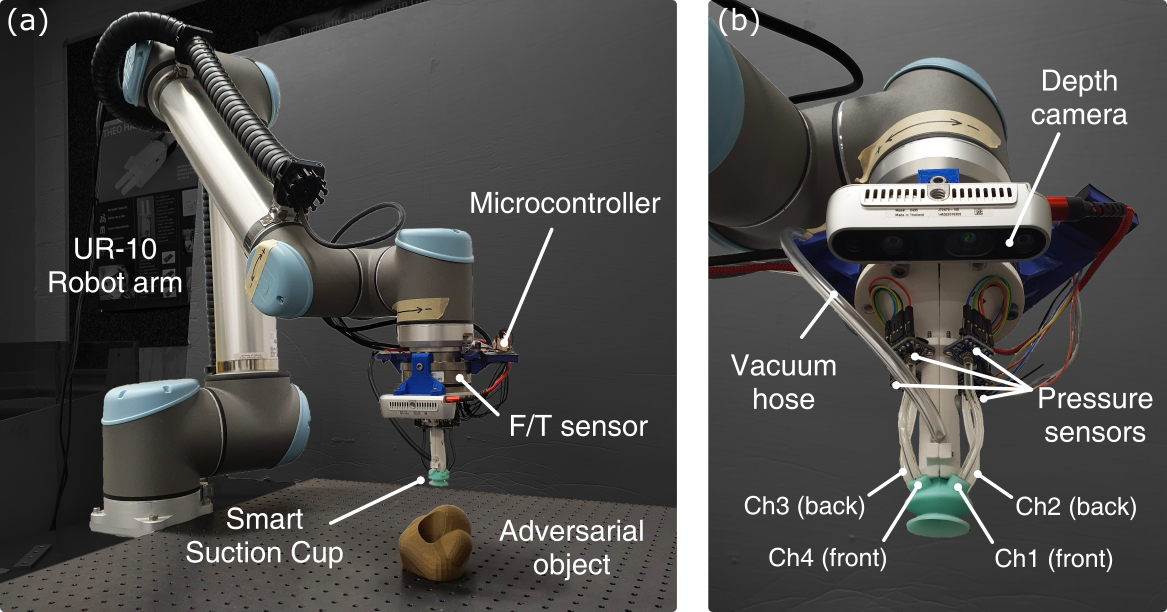}
	%\end{subfigure}
    \vspace{+1pt}
	\caption{System integration of the Smart Suction Cup. (a) the smart suction cup system integrated on UR-10 robotic arm with a 6 DOF F/T sensor and a microcontroller. (b) Close up of end-effector, including the depth camera.}
	\label{fig:cupMounting}
	\vspace{-15pt}
\end{figure}

Four ported pressure sensors (Adafruit, MPRLS Breakout, 24 bit ADC, 0.01 Pa/count with an RMS noise of 5.0 Pa) connect with the four chambers of the smart suction cup via polyurethane tubes. %For the pressure sensor connected to each internal chamber, we used a ported pressure sensor (Adafruit, MPRLS Breakout, 24bit ADC, 0.01Pa/count) that has an RMS noise of 5.0 Pa. 
%Pressure sensors are remotely connected to the suction cup via a polyurethane tube, allowing it to remain small and flexible. 
The suction cup and the pressure sensors attach to a 3D printed fixture (\cref{fig:main}a) and this fixture is attached to the wrist F/T sensor (ATI, Axia80, sampling rate 150 Hz) on the robot arm (Universal Robots, UR-10) as in \cref{fig:cupMounting}. A microcontroller (Cypress, PSoC 4000s) is fixed to the arm proximal to the load cell and communicates with the four pressure sensors via I2C at a 166.7 Hz sampling rate. 

A vacuum generator (VacMotion, VM5-NA) converts compressed building air to a vacuum source with a maximum vacuum of 85 kPa. A solenoid valve (SMC pneumatics, VQ110, On/off time = 3.5 / 2 ms), commanded by a microcontroller, regulates the compressed air as a means of moderating vacuum intensity. %\st{During haptic exploration experiments, the valve is controlled with pulse width modulation (PWM) at a frequency of 30\,Hz with 30\% duty cycle to approximate lower vacuum pressures. We chose this PWM setting considering the on-off time of the solenoid valve and the sampling rate of the pressure sensor.} 
% The generator is connected to a regulated compressed air (70kPa) through a solenoid valve (SMC pneumatics, VQ110, On/off time = 3.5 / 2ms). we used this fast switching valve to implement PWM control to the vacuum pressure for the haptic exploration experiments. 
The vacuum hose that applies suction to the cup is attached at both the suction cup vacuum connector and proximal to the load cell to reduce tube movement and subsequent F/T coupling.

The experiments are conducted on a desktop computer running Ubuntu 20.04 with a 3.00-GHz Intel Core i5-7400 quad-core CPU and an Intel HD Graphics 630 GPU. The UR-10 controller is responsible for moving the robot to the target pose, while communication between the desktop computer and the UR-10 robot uses Real-Time Data Exchange (RTDE) over a standard TCP/IP connection. We used ROS (Noetic) to collect both pressure sensor and wrist force/torque (F/T) sensor data during experiments.
An RBG-D camera (Intel, RealSense D435) is additionally mounted to the robot arm wrist such that it does not apply any wrenches on the F/T sensor. It takes photos (640x480 RGB resolution, 0.1 mm depth resolution), which are used in the bin-picking experiments.

\section{Autonomous Haptic Search}
% \section{Haptic Search Method}
\label{sec:algorithms}

% \subsection{From Pressure Signals to Motion Primitives}
% \subsection{Haptically-informed Motion Primitives}

\begin{figure}[t]
\centering
	%\vspace{-10pt}
	%\begin{subfigure}[h]{0.5\textwidth}
	%\centering
	\includegraphics[width=1\linewidth]{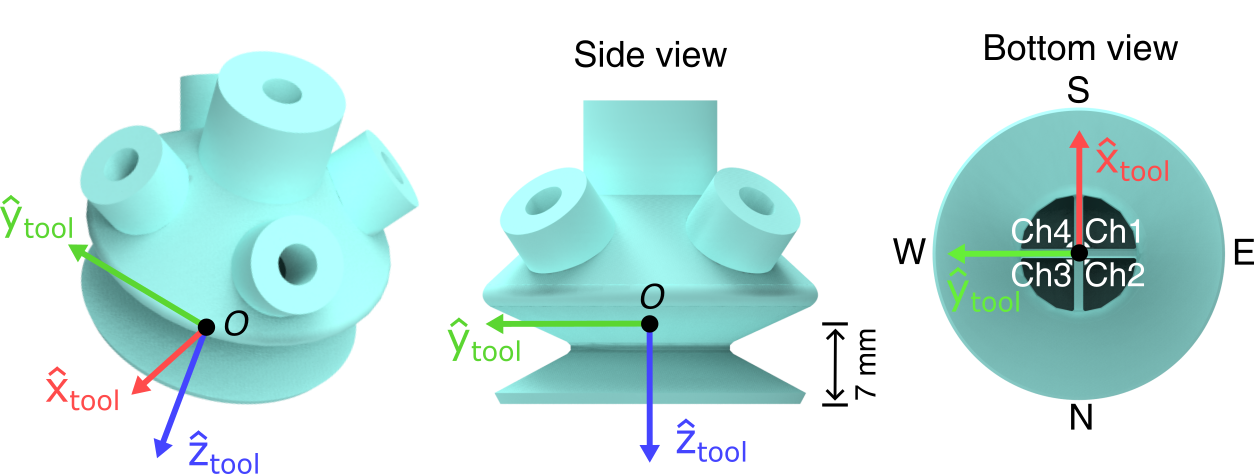}
	%\end{subfigure}
    \vspace{+2pt}
	% \caption{System integration of the smart suction cup. Left: the smart suction cup system integrated on UR-10 robotic arm with a 6 DOF F/T sensor and a microcontroller. Right: depth camera}
    \caption{The reference frame associated with the tool end is shown, including the origin point ($O$) located relative to the unloaded cup lip. The cardinal directions of the cup are oriented along the walls of the inner chamber, shown in the bottom view.}
	\label{fig:cup_axis}
	\vspace{-15pt}
\end{figure}

The control goal is to enable the robot arm to make small end-effector pose adjustments in the direction that will eventually seal the suction cup, in other words bring the vacuum pressure of all channels closer to the maximum vacuum--85kPa for the fully sealed condition. We decompose autonomous haptic search motions into three direction unit vectors defined in the tool basis, shown in Figure \ref{fig:cup_axis}: (1) \textit{lateral positioning} or translation along $\hat{v}$ in the $\hat{x}$-$\hat{y}$ plane, (2) \textit{rotational alignment} or rotation about $\hat{\omega}$ in the $\hat{x}$-$\hat{y}$ plane, and (3) \textit{axial movement} or movement along $\hat{z}$. %The magnitude of motion along the direction vectors is set by the three motion primitives with step sizes $\Delta L$, $\Delta \theta$, and $\Delta z$, respectively. 
The lateral positioning assumes partial contact of the suction cup with an object or the presence of small holes underneath the cup. The rotational alignment assumes a misalignment between the suction cup and the surface normal of the object contact point. In both situations, we assume there is significant misalignment or the existence of bottom holes, resulting in vertical leak flows as depicted in \cref{fig:cupCFD_simulation}(a). The axial movement ensures a consistent normal force, or $\hat{z}$-force, that is necessary to engage the suction with an object and maintain contact. 

% For axial movement, behavior is based only on the direct F/T sensor reading to maintain a consistent $\hat{z}$-force.
Both lateral positioning and rotational alignment search for a better grasping pose using smart suction cup pressure signals.
% However, motion primitive unit vectors used in rotational alignment and  lateral positioning rotate and slide the cup along the surface based on %. These directions are calculated by the 
% measured smart suction cup pressure signals. %To do this, we first measure the pressure differentials across the chambers, which we then use to calculate desired motion %direction vectors 
%at each time step in real time.  
%For (3) axial movement, step size is based on the F/T sensor and controlled to maintain a consistent $\hat{z}$-force. % keep the axial force within [1.5 N, 2.0 N].
To do so, pressures are first calculated for each cardinal direction by taking the average of the two chambers that correspond to that direction:\footnote{This first step aligns the cardinal points with the wall interfaces of the cup. Alternatively, one can directly assign chambers to cardinal directions, e.g., $P_E=P_{1}$; this would result in a tool basis rotation of $45^{\circ}$ about the $\hat{z}$ direction compared to our implementation.}
\begin{subequations}
    \begin{align}
        P_{E} = (P_{1} + P_{2})/2 \\
        P_{N} = (P_{2} + P_{3})/2 \\
        P_{W} = (P_{3} + P_{4})/2 \\
        P_{S} = (P_{4} + P_{1})/2 .
    \end{align}
\end{subequations}
%This averaging has the benefit of \todo{[why did you do this? does the averaging make the signal less sensitive to noise? some other reason? e.g. why did you not just average 1 and 3 to get one cardinal direction directly?].}
Pressure differentials across cardinal directions are then calculated as:
\begin{subequations}
  \begin{align}
    \Delta P_{WE} = P_{W} - P_{E} \\
    \Delta P_{NS} = P_{N} - P_{S} .
  \end{align}
\end{subequations}
Using these values, the vectors $\hat{v}$ and $\hat{\omega}$ %\todo{and scalars $\Delta L$, $\Delta \theta$, and $\Delta z$} 
are calculated at each time step, in real time at a control rate of 125Hz.

\subsection{Pressure Signal to Lateral Positioning}

%Given these two pressure differential measurements, we can calculate lateral and rotational direction vectors. 
 
The lateral direction vector, $\hat{v}$, is defined to move the suction cup towards the channels with less leakage flow, i.e., higher vacuum pressure, as follows:
\begin{subequations}
% \vspace{-15pt}

  \begin{gather}
    \vec{v} = -\Delta P_{NS} \hat{x}_{tool} + \Delta P_{WE} \hat{y}_{tool} \\
    \hat{v} = \vec{v} / ||\vec{v}|| .   
  \end{gather}
  \label{eqn:lat_direction}
%https://www.overleaf.com/project/6271646c4e8b810e71765663
% \vspace{+5pt}
% \label{eqn:P_vac_Def}
\end{subequations}
% The lateral direction vector is normalized and has components in the $\hat{x}$ and $\hat{y}$ directions of the tool frame: 
% \begin{equation}
% \hat{v} = \vec{v} / ||\vec{v}|| = [v_x, v_y, 0]^T.
% \label{eqn:lat_direction}
% \end{equation}
Then the lateral repositioning increments, $\Delta L_x$ and $\Delta L_y$, are defined as follows:
\begin{subequations}
  \begin{align}
    \Delta L_x(\hat{v}, \Delta L) = \Delta L \hat{v} \cdot \hat{x}_{tool} \\
    \Delta L_y(\hat{v}, \Delta L) = \Delta L \hat{v} \cdot \hat{y}_{tool}
    % \Delta L_x(\hat{v}, \Delta L) = \Delta L v_{x} \\
    % \Delta L_y(\hat{v}, \Delta L) = \Delta L v_{y}
  \end{align}
\end{subequations}
where $\Delta L = $ 0.5 mm, is the overall lateral positioning step size per control loop.

\subsection{Pressure Signal to Rotational Alignment}

The rotational direction vector (axis of rotation), $\hat{\omega}$, is defined to close the gap between the object and channels with high leakage flow, i.e., low vacuum pressure, as follows:
\begin{subequations}
% \vspace{-15pt}
\begin{gather}
\vec{\omega} = -\Delta P_{WE} \hat{x}_{tool} - \Delta P_{NS} \hat{y}_{tool} \\
\hat{\omega} = \vec{\omega} / ||\vec{\omega}|| = [\omega_1, \omega_2, 0]^T.
\end{gather}
% \vspace{+5pt}
% \label{eqn:P_vac_Def}
\label{eqn:rot_direction}
\end{subequations}
% \begin{equation}
% % \vspace{-15pt}
% \vec{\omega} = -\Delta P_{WE} \hat{x}_{tool} - \Delta P_{NS} \hat{y}_{tool}.
% % \vspace{+5pt}
% % \label{eqn:P_vac_Def}
% \end{equation}
% The rotational direction vector is then normalized: 
% \begin{equation}
% \hat{\omega} = \vec{\omega} / ||\vec{\omega}|| = [\omega_1, \omega_2, 0]^T.
% \label{eqn:rot_direction}
% \end{equation}
Given an overall rotational alignment step size of $\Delta \theta$ = 0.5$\degree$, the rotation matrix R is calculated as follows:
\begin{equation}
% e^{\Delta \theta \hat{\omega}^{\times}} =
R(\hat{\omega}, \Delta \theta) =  e^{\Delta \theta S(\hat{\omega})} 
% \in \mathbb{R}^{3x3} 
\in SO(3), 
\end{equation}

where $S$ is the \revision{skew-symmetric operator},
\begin{equation}
% \hat{\omega}^{\times} = S_{\omega} = \tilde{\omega} = 
S(\hat{\omega}) = 
\begin{bmatrix}
0 & 0 & \omega_2\\
0 & 0 & -\omega_1\\
-\omega_2 & \omega_1 & 0
\end{bmatrix}
\end{equation}
%\todo{Define a point in the cup that is the center of rotation in any direction. i.e. show the depth of the origin of the reference frame in Fig 6, assuming that is the point. Perhaps can define distance from the lip?.}
Rotations are applied about the axis of rotation, along $\hat{\omega}$, which is always in the $\hat{x}$-$\hat{y}$ plane and always intersects point $O$.

\subsection{Force Signal to Axial Motion}

The axial step size $\Delta L_z$, is calculated as follows,
\begin{equation}
\Delta L_z = 
\left\{
    \begin{array}{lr}
        -\Delta z, & \text{if } F_z \leq F_{z,min} = 1.5N\\
        0, & \text{if }  F_{z,min} < F_z < F_{z,max}\\
        \Delta z, & \text{if } F_z \geq F_{z,max} = 2.0N
    \end{array}
\right\}
\end{equation}
where $\Delta z =$ 0.1 mm is the axial step size per control loop.

\subsection{Composition of Motion Primitives}

To test different combinations of lateral and rotational motion in experiments, step sizes in the lateral and rotational directions are scaled as: % according to the following equations:
% \begin{equation}
% \Delta \theta_{\alpha} = \Delta \theta
% \end{equation}
% \begin{equation}
% \Delta L_{\alpha} = \Delta L (1-\alpha)
% \end{equation}
\begin{subequations}
  \begin{align}
    \Delta \theta_{\alpha} = \Delta \theta \alpha \\
    \Delta L_{\alpha} = \Delta L (1-\alpha)
  \end{align}
\end{subequations}
where $\Delta \theta_{\alpha}$ and $\Delta L_{\alpha}$ are new step sizes weighed by $\alpha$, which in turn change $\Delta L_x$, $\Delta L_y$, and $R$, producing an overall transformation matrix, $T$:
% \begin{equation}
% \begin{bmatrix}
%  &  &  & \Delta L_x (v, \Delta L_{\alpha}) \\
%  & R(\omega, \Delta \theta_{\alpha}) &  & \Delta L_y (v, \Delta L_{\alpha}) \\
%  &  &   &  \Delta L_z \\
% 0 & 0 & 0 & 1 
% \end{bmatrix} 
% \end{equation}
\begin{equation}
T = 
\left[\begin{array} {c c c | c}
     &  &  & \Delta L_x (\hat{v}, \Delta L_{\alpha}) \\
     & R(\hat{\omega}, \Delta \theta_{\alpha}) &  & \Delta L_y (\hat{v}, \Delta L_{\alpha}) \\
     &  &   &  \Delta L_z \\
    \hline
    0 & 0 & 0 & 1 
\end{array} \right]
% \in \mathbb{R}^{4x4}
\in SE(3)
\label{eqn:target_pose}
\end{equation}

If $\alpha=0$, then $\Delta \theta_{\alpha}$ = 0 and $\Delta L_{\alpha}$ = 1, which results in pure lateral positioning. If $\alpha=1$, then $\Delta \theta_{\alpha}$ = 1 and $\Delta L_{\alpha}$ = 0, which results in pure rotational alignment. For any $\alpha$, axial force control remains unchanged to ensure contact with a surface.

\section{Experimental Methods}
\label{sec:methods}

\subsection{Sensing Characterization for Haptic Search}
% \subsection{Haptic Characterization of Lateral Positioning}
% \subsection{Haptic Characterization of Rotational Alignment}

\begin{figure}[t!]
    \centering
    \includegraphics[width=0.95\linewidth]{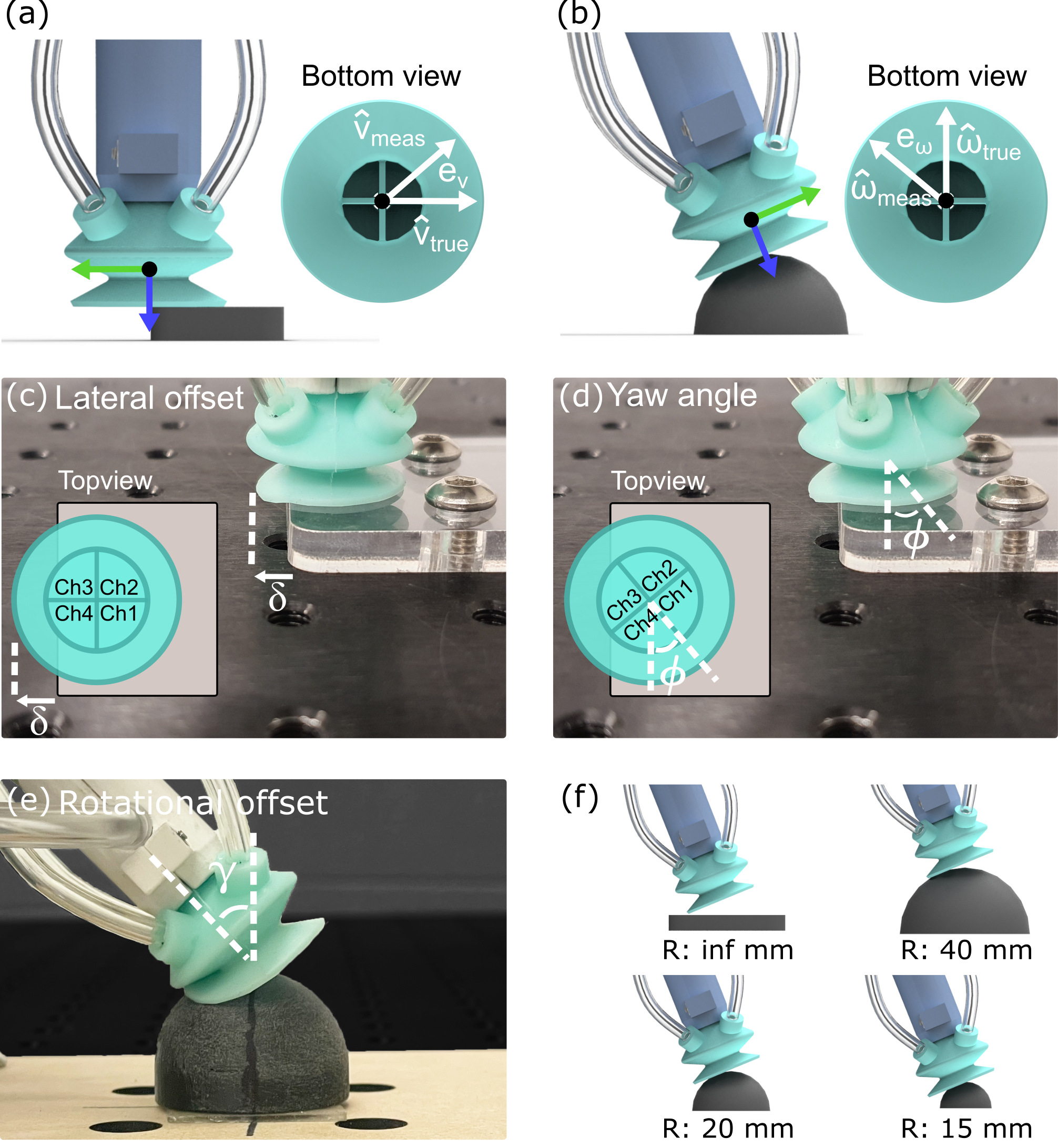}
    \vspace{1pt}
    \caption{Schematic image of direction error for (a) lateral positioning and (b) rotational alignment. (c) Experimental image of the suction cup with lateral offset, defined as the exposed lip length $\delta$, and (d) yaw angle $\phi$ about the symmetric axis of the cup. (e) Experimental image of the suction cup with a rotational offset angle $\gamma$ on a dome. (f) Four different radius domes for characterization of rotational alignment.}
    \label{fig:characterization}
    % \vspace{-5pt}
%\end{figure}
\vspace{1mm}
%\begin{figure}[b!]
    \centering
    \includegraphics[width=0.95\linewidth]{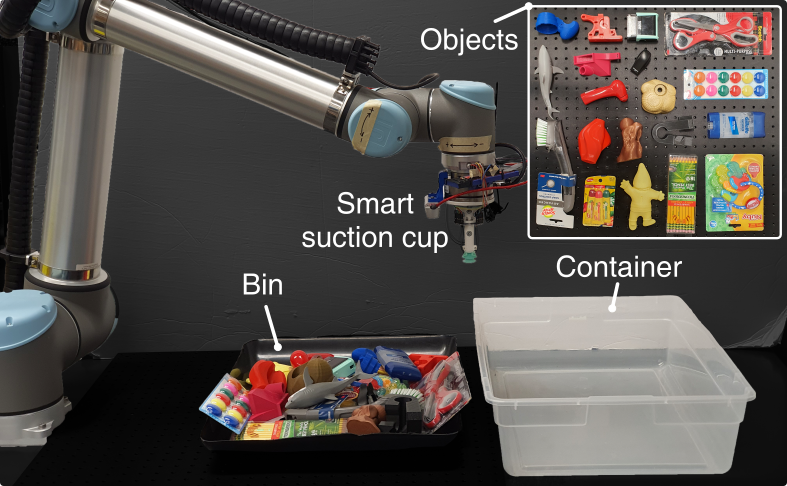}
    \vspace{3pt}
    \caption{Tabletop setup for bin picking experiments. Inset: a dataset of 19 adversarial objects, showing eight 3D printed objects, six real objects with packaging, and five real objects without a package.}
    \label{fig:Binpicking_setup}
    \vspace{-20pt}
    % \vspace{-3mm}
\end{figure}

To characterize the Smart Suction Cup sensing performance relevant for (1) lateral positioning and (2) rotational alignment, we perform two characterization experiments, one for each. We swept lateral and rotational offsets from known reference points and analyzed the resulting pressure signals. From these pressure signals in each experiment, we compute measured $\hat{v}=\hat{v}_{meas}$ and $\hat{\omega}=\hat{\omega}_{meas}$, respectively. Based on the physical experimental setups, we know the ground truth $\hat{v}_{true}$ and $\hat{\omega}_{true}$ that would move the suction cup towards a successful suction grasp with the shortest displacement. %We compare the true and measured directions using the angle between them, representing error
%Given that $v_{meas}$ and $\omega_{meas}$, the measured direction vectors, are calculated using the above equations, and that $v_{true}$ and $\omega_{true}$ are the true direction vectors, 
As shown in \cref{fig:characterization}a-b, we report direction error as the unsigned angle between the measured and true direction vectors:  %$e_v$ and $e_{\omega}$, as: % the angle between the direction vectors with the following expressions:
\begin{subequations}
  \begin{align}
e_v = \cos^{-1}(\hat{v}_{true} \cdot \hat{v}_{meas})
\label{eqn:e_vel} \\
e_{\omega} = \cos^{-1}(\hat{\omega}_{true} \cdot \hat{\omega}_{meas})
\label{eqn:e_omega}
\end{align}
\end{subequations}
for lateral positioning and rotational alignment, respectively, where $e_v$, $e_{\omega}$ $\in [0^{\circ}, 180^{\circ}]$.

% \vspace{-50pt}

\subsubsection{Lateral Positioning characterization procedure}
In the lateral haptic characterization experiments, we positioned and oriented the suction cup relative to the edge of a flat plate, as shown in \cref{fig:characterization}c-d. We define the lateral offset $\delta$ as the exposed lip length, and the orientation is parameterized by the yaw angle $\phi \in [0^{\circ}, 360^{\circ}]$ to test for asymmetry in the pressure sensor response. A yaw angle of $\phi=0^{\circ}$ corresponds to $\hat{v}_{true}=-\hat{y}_{tool}$. % \hs{is a cardinal direction facing a particular way?}.
To maintain a constant vertical distance between the flat plate and suction cup across all trials, we apply a normal force %applied on the F/T sensor at 0 mm lateral offset as 
of 1.5 N at a lateral offset of 0 mm and fix this height of the suction cup. %at that position. 
We sweep the lateral offset from 0 to 23 mm with a 1 mm increment, noting that an offset of 11.5 mm is when the point $O$ is vertically aligned with the edge of the plate,
and sweep the yaw angle from $0^{\circ}$ to $360^{\circ}$ with a $5^{\circ}$ increment. In each test pose, we %measure pressure readings for a duration of 3 seconds. To ensure consistency, we 
average the sensor data over a measurement period of 2 seconds. %, excluding the initial and final 0.5 seconds of the measurement which may vary due to the starting and stopping of flow.

\begin{figure*}[t!]
    \centering
    \includegraphics[width=1\linewidth]{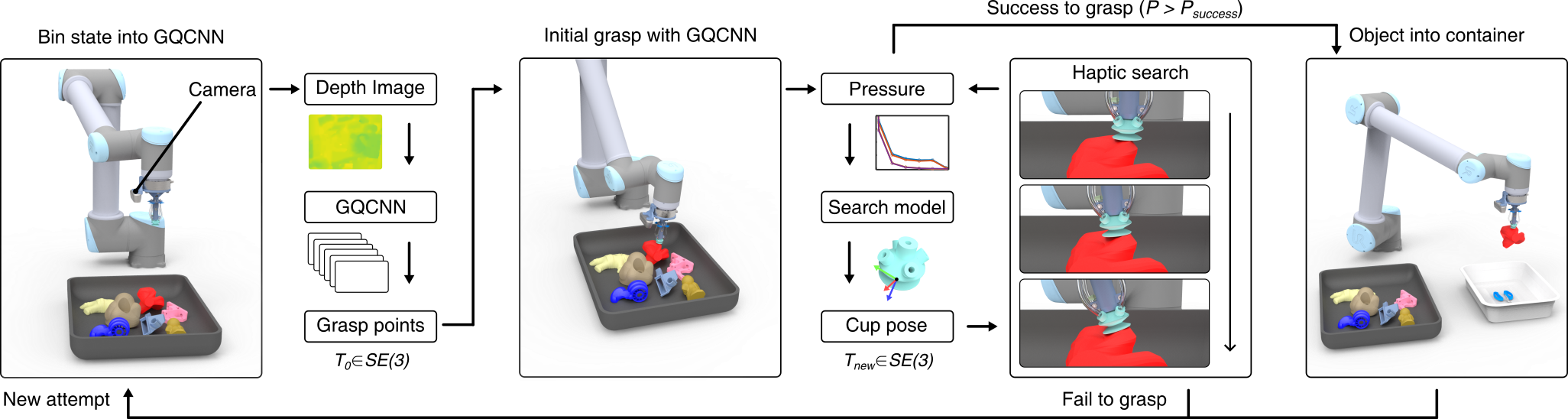}
    % \vspace{1pt}
    \vspace{-9pt}
    \caption{Flow chart of robotic behavior during bin picking experiments.\protect\footnotemark} %\hs{remove the legend on and axis labels on the "Pressure" plot so readers don't think they shouls get anything from this. Also remove symbols on the "cup pose" image.}}
    \label{fig:Binpicking_flowchart}
    \vspace{-15pt}
\end{figure*}

\footnotetext{\revision{Visual renderings are used for illustrative purposes only. All characterization and bin-picking experiments were done with physical hardware, and not in simulation.}}

\subsubsection{Rotational Alignment characterization procedure}
In rotational haptic characterization, the suction cup was placed on and oriented relative to a sphere, as in \cref{fig:characterization}e, such that the point $O$ is vertically aligned with the highest point of the dome. We define the rotational offset $\gamma$ as the angle between the true surface normal at this highest point (vertically upward) and $-\hat{z}_{tool}$. Domes with different diameters (15 mm, 20 mm, 40 mm, and flat plate) are selected, as in \cref{fig:characterization}f, with the 15mm radius dome representing the smallest sphere that the suction cup can grasp in this study. In this experimental setup, $\hat{\omega}_{true}=\hat{x}_{tool}$. 
%\revision{We designate a position of $O$ as the pivot at $\gamma=0$ for applying rotation offsets by moving the robot along its normal direction until it reaches the target normal force of 1.5 $\pm$ 0.1N. We found a tolerance of 0.1 N did not result in substantial vibrations.}
\revision{To initialize an experiment, we use force control to reach a target 1.5$\pm$0.1 N normal load\footnote{\revision{This type of force control to an exact value often leads to system vibrations. For the parameters used in our controller, with the tolerance of $\pm$0.1 N and control rate of 125 Hz, we did not observe substantial vibrations}.}, with $\gamma=0$. We record the position of $O$ in space at this moment, and then pivot about it while regulating the force along $\hat{z}_{tool}$.}
%We use force control to reach a target 1.5 N load with $\gamma=0$ %on the peak of the dome 
%and set this position of $O$ as the pivot for applying rotational offsets.
We sweep $\gamma$ from 45$^{\circ}$ to 0$^{\circ}$ with 1$^{\circ}$ steps. At each offset, we average pressure measurements for 2 seconds of steady state readings. %, and also to ensure consistency, we ignore the initial and final 0.5 seconds. 
%Vacuum pressure at each offset is calculated as the mean within the 2-second window. 
%Both direction error $e_{\omega}$ and its mean at each offset are plotted to illustrate error distribution.

%\subsection{Direction Error}

%\begin{figure}[tbp!]
%\centering
	%\vspace{-10pt}
	%\begin{subfigure}[h]{0.5\textwidth}
	%\centering
%	\includegraphics[width=0.9\linewidth]{./Figures/direction_errors.png}
	%\end{subfigure}
%    \vspace{+10pt}
	% \caption{System integration of the smart suction cup. Left: the smart suction cup system integrated on UR-10 robotic arm with a 6 DOF F/T sensor and a microcontroller. Right: depth camera}
%    \caption{\hs{this figure should be improves. I don't see much benefit to the top images, as they seem redundant to Fig 7. Also the symbols on the bottom get lost. I recommend just defining the angle $e_v$ and $e_{\omega}$ in the text as the angles between two vectors that lie in the x-y tool frame. The images are helpful for us, but I don't think critical for the paper.}}
%	\label{fig:direction_error}
%	\vspace{-15pt}
%\end{figure}

\begin{figure*}[b!]
\centering
	\vspace{-10pt}
    % \vspace{-15pt}
	%\begin{subfigure}[h]{0.5\textwidth}
	%\centering
	\includegraphics[width=0.95\linewidth]{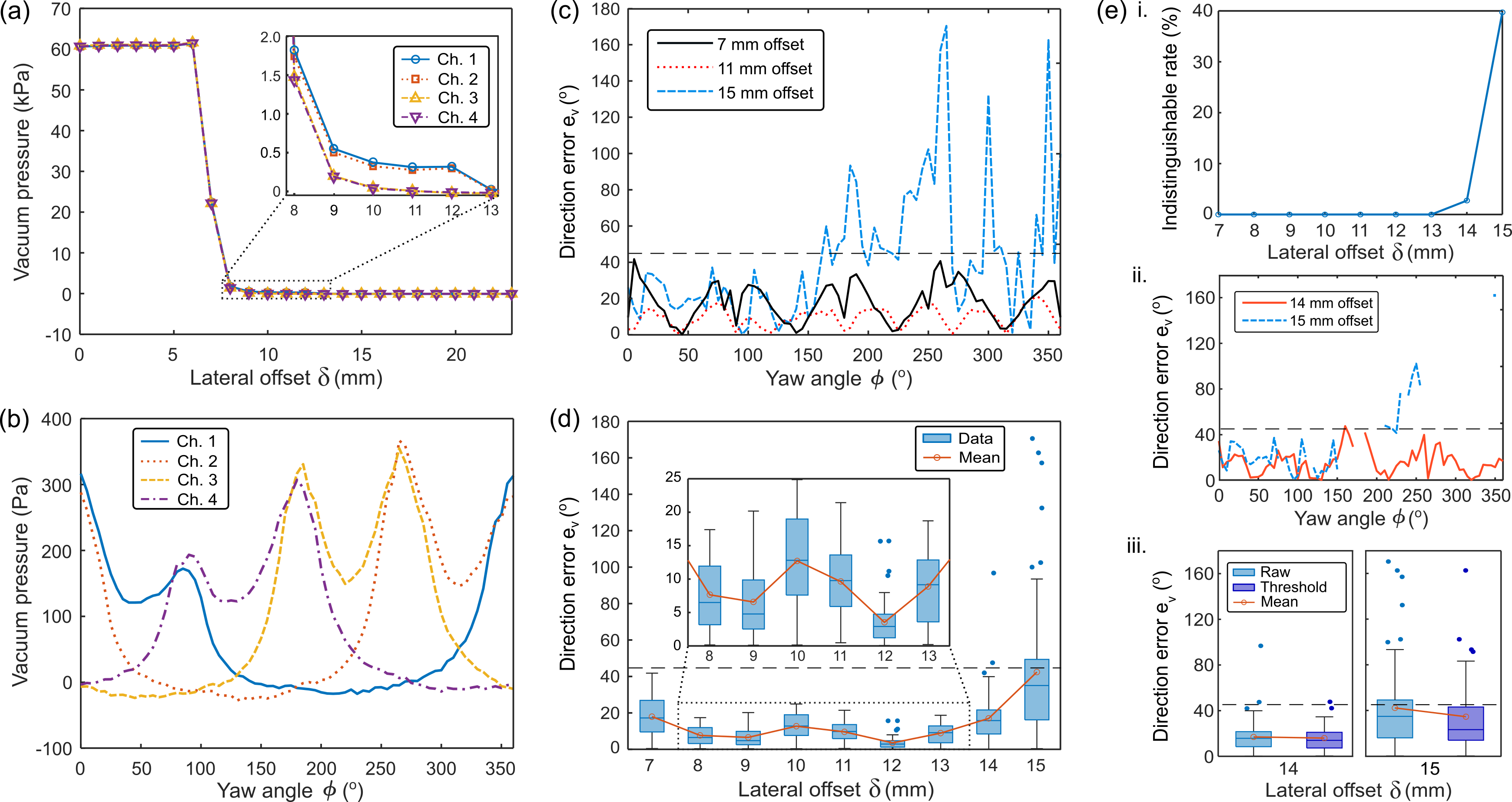}
	%\end{subfigure}
    \vspace{1pt}
	\caption{The pressure sensor readings for a sweep of lateral offset, $\delta$, and yaw angle, $\phi$, for the lateral positioning. (a) Vacuum pressure for a sweep of lateral offset from 0 to 23 mm at 0$\degree$ yaw angle. (b) Vacuum pressure reading for a sweep of yaw angle from 0$\degree$ to 360$\degree$ at the center of the suction cup by averaging pressure reading at 11 and 12 mm lateral offset. 
 %(c) The estimated yaw angle of various lateral offsets, including the dashed line shows the upper and lower boundary of 45$\degree$ from the yaw angle. 
 (c) The direction error of various lateral offset.
 (d) Direction error data and mean. (e) Results from thresholding pressure readings by 10 Pa. i. The indistinguishable rate for a sweep of lateral offset. 
 %ii. The estimated yaw angle at 14 mm and 15 mm lateral offset without indistinguishable data by thresholding pressure readings.
 ii. The direction error at 14 mm and 15 mm lateral offset without indistinguishable data by thresholding pressure readings.
 iii. Direction error data and mean before and after thresholding pressure readings at 14 mm and 15 mm lateral offset. Dashed lines in the figure represent 45$\degree$.}% \hs{aces and text are smallaer in (e) than the other plots. Please increase to match.}}
	\label{fig:lateral}
	% \vspace{-15pt}
    \vspace{-5pt}
\end{figure*}

\subsection{Bin-picking}
\label{sec:binpickingMethods}
% describe experiment protocol: baseline (brownian motion), each controller (weights), procedure, # of trials, termination conditions.

We set up a bin-picking task similar to that of \cite{mahler2019learning} to evaluate the functional performance of the proposed haptic search algorithms. % and compare them to non-haptic approaches.
The robot system was programmed to pick objects up from a bin and transport them to a designated container, as shown in \cref{fig:Binpicking_setup}. For a given trial, the robot was first set with a particular controller. The system was then presented with 19 adversarial objects in a bin. Five of the objects were 3D-printed objects taken directly from the list of Adversarial objects from \cite{mahler2018dex}. Eight of them were taken directly from the Level 3 object set in \cite{mahler2019learning}, which includes both 3D-printed and commercial objects. The rest of the objects were picked based on difficulty for a vision-based planner, specifically adversarial objects with imperceptible features like transparency, reflectivity, and small surface features. %Adversarial objects were picked to showcase the utility of the Smart Suction Cup and its haptic search capabilities.
% \hs{[This is where I would recommend clarifying how/why these particular objects are selected (both why adversarial -- citation of taking these from a collection of prior works? -- and why we want adversarial objects -- because haptic search not useful if the grasp succeeds at first].}

Before the start of each trial, the operator placed the complete set of objects into the bin by first shaking them loosely in the container, inverting that container to drop them into the bin, and manually adjusting objects only to ensure that they were below the rim of the bin. The robot then continuously attempted to perform the pick-and-place task until an end-trial condition was met, and the number of successfully grasped objects was recorded.
%Each trial is considered complete after all objects are successfully removed, 
In each trial, 57 attempts (three times the number of objects) were performed, and the trial stopped when 10 consecutive grasp attempts failed or
 no feasible grasping points remained available. We conducted five bin-picking trials for each %haptic and non-haptic 
 tested control method. 
%We define an experiment for a given haptic/non-haptic control method as comprising five independent bin-picking trials. 

%Prior to each experiment, we calibrate the position and orientation of the camera relative to the suction cup using an ArUco board.

The process for each trial is shown in \cref{fig:Binpicking_flowchart}.
%In this study, the robot attempts an iterative process to extract objects from the bin. 
On each grasp attempt within a given trial, a point cloud of the bin state with objects is inputted into the Grasp Quality Convolutional Neural Networks (GQCNN) \cite{satish2019policy} to generate 
% \hs{up to} 
30 grasp point candidates with a grasp quality value ranging from 0 to 1 and corresponding suction cup pose. Among the candidates, the pose with the highest quality value and no previous failures is attempted.\footnote{We implement a simple memory system to avoid repeated failures at the same grasp point. When a grasp is unsuccessful, the grasp point is stored and any points within 3 cm of previous failure points are considered non-feasible. The system stores up to three previous failures and is reset when the suction cup successfully grasps an object.} Note that we have not re-trained this algorithm for our particular robot system or object set. %The system selects the grasp point with the highest grasp quality value among the feasible candidate points.
The robot approaches the selected grasp point with a 15 mm offset in the estimated surface normal direction. Then, it approaches the surface along the estimated normal until normal force reaches 1.5 N. The suction cup then initiates vacuum suction and checks the vacuum pressure of all channels to determine whether it has successfully grasped an object. We define a grasp success if the mean vacuum pressure is greater than $P_{success}=$15 kPa, equivalent to holding $\sim$3\revision{5}0\,g with our suction cup. 
\revision{This estimate assumes the seal ring diameter is at the midpoint of the suction cup lip, or 17 mm across. The heaviest object lifted in experiments weighs less than 200\,g, providing a safety margin of at least 150\,g. }%We select this value with a safety margin, taking into consideration that objects used in the experiment weigh less than 200 g.}

If a successful grasp is not detected after the initial grasp attempt with GQCNN, %the suction cup fails to grasp an object in the first place, 
then the robot starts its specified search strategy to adjust the cup pose. During this search phase, a grasp is considered a failure if the suction cup moves away from the initial grasp point by more than 3 cm, rotates by more than 45$\degree$ from the initial pose, or if the search time exceeds 15 seconds.\footnote{This maximum search time of 15 seconds was selected after preliminary experiments yielded diminishing grasp success after this time frame. In applications where speed is important, it would be impractical to search for an un-ending amount of time without a successful grasp.} If the robot fails to grasp an object, it returns to the initial position and starts a new attempt. 
However, if at any point during the search procedure a successful grasp is detected, the robot then attempts to lift and move the object. Grasp failure is recorded if the object is dropped prior to the intentional release of the object into the container.

We evaluate eight total experiments: six with different haptic searching methods and two experimental controls.  %Of the search methods, six haptic search methods that utilize the pressure readings of the Smart Suction Cup and two non-haptic methods.  
We implement five haptic search strategies by modifying the value of $\alpha$ from 0 to 1 in increments of 0.25. Specifically, we denote the values of $\alpha_1$, $\alpha_2$, $\alpha_3$, $\alpha_4$, and $\alpha_5$ as an $\alpha$ of 0, 0.25, 0.5, 0.75, and 1, respectively. Also, we include a haptic search strategy which alternates a weight value between $\alpha_1$ and $\alpha_5$ every 0.5 s, denoted as $\alpha_{1\&5}$ in order to test the decoupling of motion between lateral positioning and rotational alignment. %\todo{[why did you choose this one?]}.
The first control condition is the application of GQCNN without any additional search method applied. As another experimental control case, we conduct a random search with Brownian motion (BM), or Weiner process, in the lateral direction; the lateral scalar step sizes in \cref{eqn:target_pose} , $\Delta L_x$ and $\Delta L_y$,
% \begin{subequations}
%   \begin{align}
%     \Delta L_x = \jp{need to add from the code} \\
%     \Delta L_y = \jp{need to add from the code}
%   \end{align}
% \end{subequations}
are chosen to make the standard deviation of the distance to be 3 cm from the initial grasp point after 15s of searching time. %This allows us to understand if it is the presence of motion after initial GQCNN, or the addition of haptic feedback from pressure sensing, that enables improved grasp performance.
%During this search phase, a grasp is considered a failure if the suction cup moves away from the initial grasp point by more than 3 cm, rotates by more than 45 degrees from the initial pose, or if the search time exceeds 15 seconds. If the robot fails to grasp an object, it returns to the initial position and starts a new attempt. 
%We implement a simple memory system to avoid repeated failures at the same grasp point. When a grasp is unsuccessful during the search phase, the grasp point is stored, and any points within 3 cm of previous failure points are considered non-feasible. The system stores up to three previous failures and is reset when the suction cup successfully grasps an object.

% \begin{algorithm}
% \caption{An algorithm with caption}\label{alg:cap}
% \begin{algorithmic}

% \State Take depth image.
% \State GQ-CNN determines best candidate grasp point.
% \State Robot arm goes to grasp point and attempts grasp.

% \If{P > P_{threshold}}
%     \State Attempt to pick up object and drop in bin
% \Else{}
%     \While{Autonomous haptic search}
%     \If{}
%     \If{t > 15s and no grasp}
%         \State End haptic search
%     \EndIf
% \EndIf

% \Require $n \geq 0$
% \Ensure $y = x^n$
% \State $y \gets 1$
% \State $X \gets x$
% \State $N \gets n$
% \While{$N \neq 0$}
%     \If{$N$ is even}
%         \State $X \gets X \times X$
%         \State $N \gets \frac{N}{2}$  \Comment{This is a comment}
%     \ElsIf{$N$ is odd}
%         \State $y \gets y \times X$
%         \State $N \gets N - 1$
% \EndIf
% \EndWhile

% \end{algorithmic}
% \end{algorithm}

\begin{figure*}[b!]
    \centering
	% \vspace{-5pt}
	%\begin{subfigure}[h]{0.5\textwidth}
	%\centering
	% \includegraphics[width=0.9\linewidth]{./Figures/1_main_ver5.pdf}
	\includegraphics[width=1.0\linewidth]{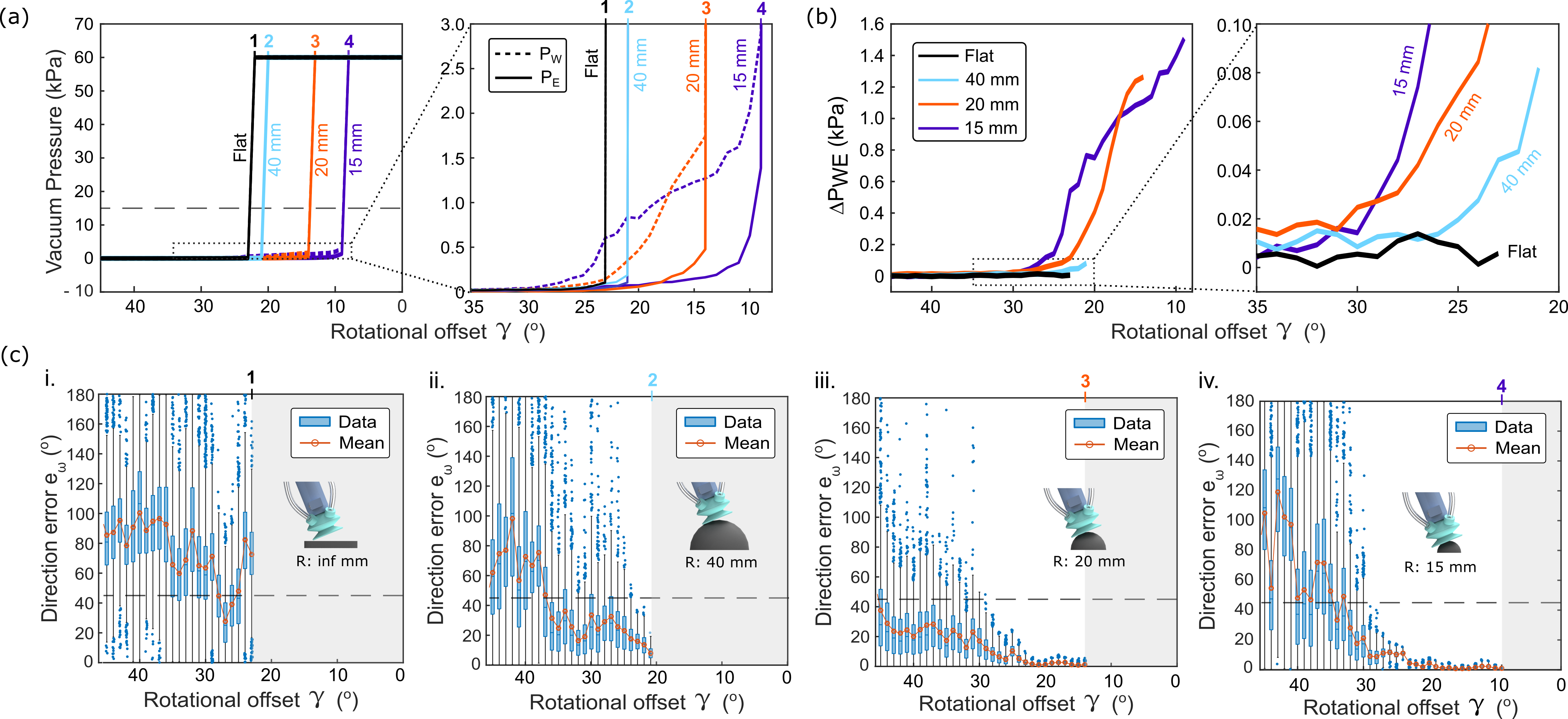}
 % {./Figures/test_rotCharac.png}
    
	%\end{subfigure}
    \vspace{10pt}
    %\vspace{+1pt}
    \caption{Vacuum pressure and pressure differential measurements for a sweep of rotational offsets $\gamma$, and direction error, $e_{\omega}$, for four different objects - a flat plate and spheres with 40 mm, 20 mm, and 15 mm radii. (a) Vacuum pressures for $\gamma \in $ [45$\degree$, 0$\degree$]. Pressure \revision{increases} sharply at different \revision{critical offset angles} as the vacuum seal\revision{s on the surface, points numbered 1-4}. \revision{Before sealing occurs,} differences between \revision{$P_W$ and $P_E$ are visible, especially for the 20 mm and 15 mm objects}. (b) Pressure differential between west and east chambers for each curved surface. %Pressure differential drops to zero at different rotational offsets depending on the object's curvature. 
    Differential signals \revision{rise faster for high curvature objects.} %the soonest for the flat plate, while for the 20mm dome, the signal drops below 10 Pa at a higher offset. 
    (c) i-iv. Direction error data and mean for the four objects. Included is the 45$\degree$ direction error boundary line. \revision{The shaded regions indicate the rotational offsets at which the suction cup passively grasps the object, smaller than the critical offset angle. Direction error past 90$\degree$} corresponds to motion perpendicular to the true desired direction.}
    % \hs{I think there are ways to streamline this data representation. The top could be condensed into two subplots, one with all full-scale plots and a second with all the "inset" date on it. I like the bottom plots as they are, but the text is too small -- it should be at least as big as the caption text.}
    % The pressure sensor readings for a sweep of rotational offset, $\theta$. The cup starts at a rotational offset of 40° and rotates towards a vertical orientation until a suction grasp is made. While rotating slowly, the cup moves axially to keep a constant axial force of 1.5N. For a ∅30mm dome, suction grasp occurs at $\theta_{grasp}$. This offset is also how tolerant the suction cup is to misalignment to the surface normal. $\Delta P$, the difference between the right and left pressure chambers, is the pressure differential that informs the direction of motion. $\theta_{recover}$ is the minimum orientation offset required to induce a meaningful $\Delta P$ and therefore direction of motion, shown here by directionality.}
	\label{fig:rotational}
	\vspace{-5pt}
\end{figure*}

\section{Results}
\label{sec:Result}

\subsection{Lateral Positioning sensor characterization}%\jp{Lateral recovery}}

The characterization results of lateral positioning are presented in \cref{fig:lateral}. 
In \cref{fig:lateral}a, the vacuum pressures from all channels are shown as lateral offset changes while yaw angle is held constant at $\phi=0\degree$. All four channels remain over 60 kPa until the lateral offset reaches 6 mm; at these offsets\revision{, less than 7 mm}, the suction cup seals completely with the plate \revision{and no haptic search is needed to grasp successfully}. Note that the entire lip of this cup design does not necessarily need to be in full contact to generate a seal. Vacuum pressures decrease starting from a 7 mm lateral offset. 
The figure inset shows the region of offsets in which notable pressure differences exist between different chambers. Between 16 mm and 23 mm lateral offset, pressure readings remain at 0 kPa across all chambers. It is therefore expected that directional signals will be most informative between 7 and 15 mm of offset.

To demonstrate how the pressure readings vary with the yaw angle, we vary $\phi$ from 0$\degree$ to 360$\degree$ with the edge of the plate located at the center of the suction cup (11.5 mm offset); we average the pressure readings at 11 mm and 12 mm lateral offset to estimate this cup alignment. As shown in \cref{fig:lateral}b, the vacuum pressures in each chamber vary periodically with the change in yaw angle. At 0$\degree$ yaw angle, chambers 1 and 2 overlap with the plate, showing higher vacuum pressure than the pressures from chambers 3 and 4. At every 90$\degree$ of yaw angle, two chambers seal on the plate's surface, causing vacuum pressures to show peaks of two chambers. Given the chamber geometry of the cup, there will be higher overall vacuum pressure applied to the cup when more of the 4 chambers become sealed. This explains why we see two peaks per chamber, rather than just one, as the two adjacent chambers simultaneously break seal between local maxima.
%As the smart suction cup contains four channels, covering more chambers leads to higher vacuum pressure. As seen in \cref{fig:lateral}b, covering two chambers generates higher overall vacuum pressure than covering only one chamber. For example, at a yaw angle of 0$\degree$, Ch. 1 shows higher vacuum pressure than at 45$\degree$, where only Ch. 1 would be in the sealing position. The suction cup response is reasonably symmetrical, leading to periodic oscillation in the pressure readings during the yaw angle sweep.
Variability between chambers is also seen, for example the maxima at $\phi=90^{\circ}$ is smaller than the others.
Small variation could be caused by fabrication and assembly, as well as compliance in the suction cup, leading to asymmetric buckling deflections of the internal chamber dividers upon contact, as observed in prior work \cite{huh2021}. %Therefore, even though the suction cup's center may be located at the plate boundary, the chambers may not always be in an ideal position, resulting in direction errors.
Regardless of chamber-to-chamber interaction and nonidealities, at each tested yaw orientation the 4 chambers provide a unique combination of readings to support the estimation of the $\phi$ state.

In order to understand the interpretation of these signals in our control algorithm, across both $\delta$ and $\phi$, we visualize direction errors $e_v$ with pressure sensor readings using \cref{eqn:e_vel} in \cref{fig:lateral}c-d. Direction errors from lateral offsets between 7 mm and 15 mm with \revision{4} mm increments are shown in \cref{fig:lateral}c. %The red line shows the true yaw angle, and the dashed lines shows $\pm 45\degree$ from the true yaw angle. 
The $45\degree$ boundary indicates the directions that would enable faster haptic search for a better grasping point, by moving the cup towards the plate at a rate faster than along the edge of the plate. \revision{At 7 mm and 11 mm lateral offset,}  the direction errors show that all data is below the $45\degree$ boundary line. At a 15 mm lateral offset, some errors go above the boundary. The result shows that direction errors have a cyclic pattern every 45$\degree$, reflecting the internal wall structure of the suction cup with four chambers. %To visualize the errors of the estimated yaw angle across $\delta$ with 1 mm spatial resolution, we report the direction error $e_v$ from \cref{eqn:e_vel} in \cref{fig:lateral}d for all trials between 7 and 15 mm offset.

In \cref{fig:characterization}d, we report the direction error for all trials between 7 and 15 mm offset. Each lateral offset has 73 data points, where we sweep yaw angles from 0$\degree$ to 360$\degree$ with a 5$\degree$ increment. The result shows box plots with the means of the data. No data exceeds the $45\degree$ boundary from 7 mm to 13 mm lateral offset. However, within this range, error is greatest at 7 mm. \revision{It makes sense that direction error increases as the offset approaches 6 mm, as the suction cup becomes fully sealed and flow stops altogether.} %which 
%From the direction errors between the start of exposing chambers and the center of the suction cup, which have a lateral offset range of 7 mm to 12 mm, the results show that the direction error at 7 mm had the highest mean error. 
%may be attributed to leakage flow
\revision{For the 7 mm case}, as demonstrated in \cref{fig:cupCFD_simulation}d-f\revision{,} flow can become predominantly horizontal at the transition to the fully-sealed state, %. This flow component causes the opposite chamber to have the lowest vacuum pressure, 
thereby decreasing the pressure difference between the exposed and covered chambers. %As a result, the opposite chamber experiences the lower vacuum pressure, reducing the pressure difference between exposed and covered chambers. 
At both 14 and 15 mm lateral offset, where %once again 
pressure differences become small, several data points show error over $45\degree$, yet the mean of the direction error remains below this threshold.

At large offsets, greater than 15 mm, the pressure approaches 0 Pa and $e_v$ increases further, meaning that all four channels are open and not forming effective differential pressures. We therefore apply a threshold condition of 10 Pa during controller implementation such that, when all chambers are below this level, the direction estimate is set to $\hat{0}$ (no motion). The rate at which this condition is met, which we call the indistinguishable rate, at different lateral offsets is shown in \cref{fig:lateral}e i. The indistinguishable rates from thresholding are found to be 2.7\% and 39.73\% at 14 mm and 15 mm lateral offset, respectively, while no data are indistinguishable between 7 mm and 13 mm lateral offset. \cref{fig:lateral}e ii shows the corresponding result of direction errors $e_v$ at 14 mm and 15 mm lateral offset, where the indistinguishable data points are eliminated. \cref{fig:lateral}e iii shows the change in $e_v$ resulting from the threshold condition. Before thresholding, the mean of direction error at 14 mm lateral offset is 16.99$\degree$, which decreases to 15.70$\degree$ after thresholding. At a 15 mm lateral offset, the mean direction error changes from 42.42$\degree$ to 34.71$\degree$.
In practice, motion will only occur when at least one channel measures a degree of flow restriction -- if there is no measurable suction contact the cup will remain stationary.

%In our experiments, we observed that the suction cup begins to expose the chamber after a 6 mm lateral offset, as depicted in \cref{fig:lateral}a.  

\begin{figure*}[b!]
    \centering
    \vspace{-10pt}
    \includegraphics[width=1\linewidth]{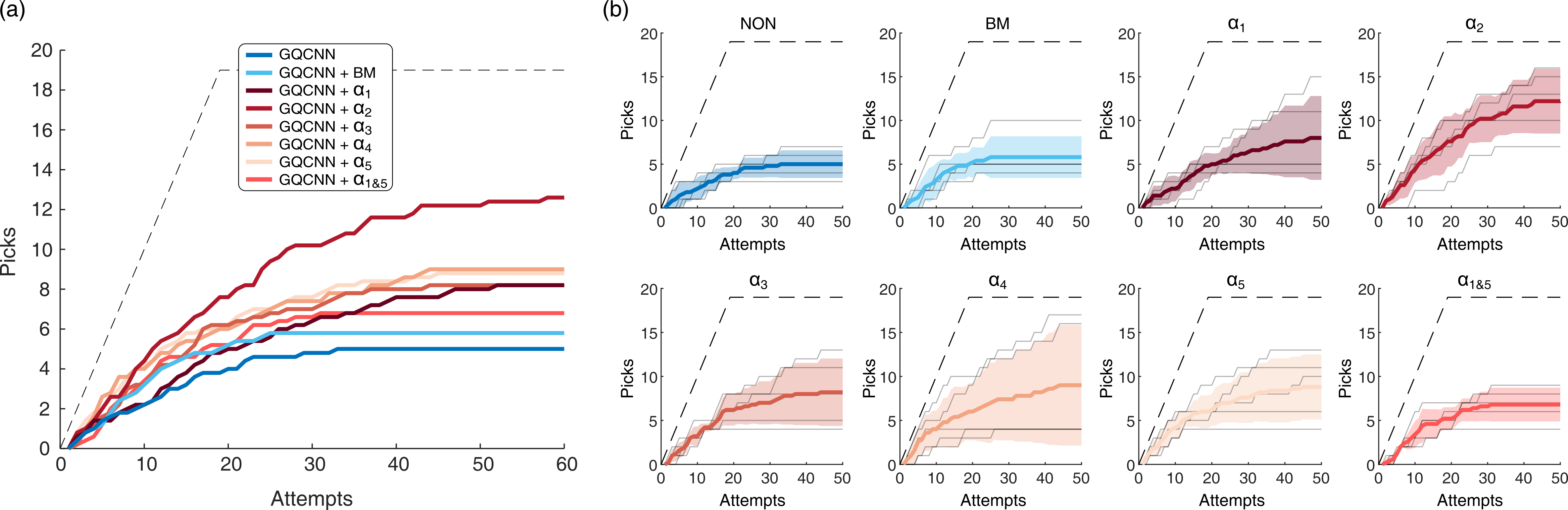}
    \vspace{-10pt}
    \caption{Results of bin picking experiments. (a) The average number of successful picks across all grasping methods. (b) The individual results for each grasping strategy, with solid colored lines indicating the average and colored areas representing the standard deviation. The grey lines within each grasping method indicate the results of individual trials. As a reference, a dashed black line is used to represent the optimal performance, which is defined as successfully picking every attempt in the bin until it is completely empty.} %\hs{(a) is beautiful though the subtext of the alpha's as hard to see. please increase legend text size. (b) text is too small, increase to match the caption size. The plots can remain small, just increase the text.}}
    \label{fig:Binpicking_result}
    \vspace{-5pt}
\end{figure*}

\subsection{Rotational Alignment sensor characterization}
The characterization results of rotational alignment are presented in \cref{fig:rotational}. Since the suction starts without a seal, we read the plot \revision{with decreasing rotational offset} from \revision{left to right}. The cup initially starts at $\gamma=45^{\circ}$ and all channels read close to 0 kPa. The critical rotational offset, where the vacuum seal is formed, is seen by a rapid increase in the vacuum pressure (\cref{fig:rotational}a). This critical offset angle becomes smaller as the radius of the dome decreases, indicating that smaller radius domes require more precise alignment with the surface normal to successfully grasp.  
% This critical angle varies depending on the dome radius, after 8$\degree$, 13$\degree$, 20$\degree$, and 22$\degree$ for domes with radii of 15 mm, 20 mm, 40 mm, and on a flat plate, respectively. %Beyond the first critical offset, pressure differences still exist, with their range also varying depending on the dome radius, i.e. surface curvature. 
At rotational offsets \revision{smaller} than the critical offset angle, the vacuum pressures are consistently near 60 kPa %\todo{over 15 kPa} 
across all channels. %The first critical rotational offset, where the vacuum seal is compromised, varies depending on the dome radius. Specifically, the vacuum seal is compromised after 8$\degree$, 13$\degree$, 20$\degree$, and 22$\degree$ for domes with radii of 15mm, 20mm, 40mm, and on a flat plate, respectively. Beyond the first critical offset, pressure differences still exist, with their range also varying depending on the dome radius, i.e. surface curvature. 
% \cref{fig:rotational}a inset illustrates the offsets at which pressure differentials become indistinguishable. 
The control condition for successful grasping, $P>P_{success}$, is shown as the horizontal dashed line.
The region of interest for haptic search occurs when there is the presence of pressure differentials within the cup, detailed in the figure, comparing $P_W$ and $P_E$. 
The difference is directly plotted as $\Delta P_{WE}$ in \cref{fig:rotational}b after removing data points where $P>P_{success}$. For the 15 and 20 mm radius domes, signals rise as high as 1.5 and 1.2 kPa, respectively, over larger rotational offset ranges than the 40 mm dome or flat plate. %It also appears that the flat plate \revision{produces $\Delta P_{WE}$ values that cannot be easily differentiated from zero.} 
\revision{The pressure differential for the flat plate in particular never even reaches a $\Delta P_{WE}$ of 20 Pa, because the compliant lip rapidly deforms and pulls itself into the surface before substantial differential flows can occur inside the cup due to chamber occlusion.}
%\revision{For the flat plate, the suction cup reaches the critical angle offset before the pressure differential signal can increase, as seen by the $\Delta P_{WE}$ values never even reaching 20 Pa.}
We therefore expect tactile sensing to provide more useful prediction of $\hat{\omega}$ on higher curvature objects, where \revision{smaller domes can better occlude chambers before the critical angle is reached} and more careful alignment with surface normal is required. %\hs{There can be either a rapid or gradual onset of sealing when approaching this critical angle, depending on object curvature and the ability of the lip to passively conform to it. }
%the signal-to-noise ratio is better at offsets of less than $30^{\circ}$.

As shown in \cref{fig:rotational}c, the test results indicate that the direction error ($e_\omega$, Eqn. \ref{eqn:e_omega}) is lower for objects with smaller radii.
Each subplot i-iv represents a trial on a different object and data for which $P>P_{success}$ is omitted. 
When we add a dashed boundary line of $45^{\circ}$, similar to in lateral search characterization, we see that errors consistently drop below $45^{\circ}$ at rotational offsets %occur at
of 30$\degree$, 31$\degree$, 23$\degree$ for domes with radii of 15 mm, 20 mm, and 40 mm, respectively. \revision{On the other hand, the flat plate error} does not fall below this threshold consistently on the flat plate \revision{because $\Delta P_{WE}$ remains small up to the critical angle}. %On 
The smaller radii objects (R=15\,mm and 20\,mm) show \revision{the most} accurate prediction\revision{s} ($e_{\omega}<10\degree$) close to the critical rotational offset. This result suggests that the proposed haptic search method can successfully grasp objects with small critical offset angles (e.g., 8$\degree$ in R=15\,mm object), even with high visual perception error of surface normal up to 30$\degree$.

\subsection{Bin-picking}
% \begin{figure}[h]
%     \centering
%     \includegraphics[width=1\linewidth]{Figures/Bin_picking_average.png}
%     \vspace{10pt}
%     \caption{Results of bin picking experiments}
%     \label{fig:Binpicking_result_average}
% \end{figure}

We evaluate the bin-picking test conditions defined in \cref{sec:binpickingMethods}, with results shown in %against two baselines in five independent trials 
\cref{fig:Binpicking_result}. 
The picks-per-attempts mean average from across five independent trials for each condition is reported in \cref{fig:Binpicking_result}a; the six haptic search conditions are in shades of red while the the two experimental control cases are in shades of blue. All trials are reported for each test condition experiment in \cref{fig:Binpicking_result}b. The dashed lines on all plots indicate the ideal case where every grasp attempt is successful without any failures. 

The control case ``GQCNN'' or ``NON,'' which has no search phase, %to guide the robot system in picking objects from bins. The results 
shows an average of 5 $\pm$ 1.58 successful picks. This means the robot system was able to successfully pick-and-place these objects from the bins without any haptic search assistance. The control case ``GQCNN + BM,'' which includes random Brownian motions in lateral direction during the search phase, results in an average of 5.8 $\pm$ 2.39 successful picks. This shows that the introduction of non-haptically-driven motion after the initial grasp attempt can provide minor improvements. Comparing the two control cases with this ideal performance, we see the difficultly of the selected adversarial pick-and-place task. Of the two control cases, we propose that it is more appropriate to compare haptically-driven results with the ``GQCNN + BM'' control case because it represents baseline benefits from the presence of a search phase. %which was used as a fair comparison for the proposed haptic search methods in terms of search time. 

\begin{figure*}[b!]
    \centering
    \includegraphics[width=1\linewidth]{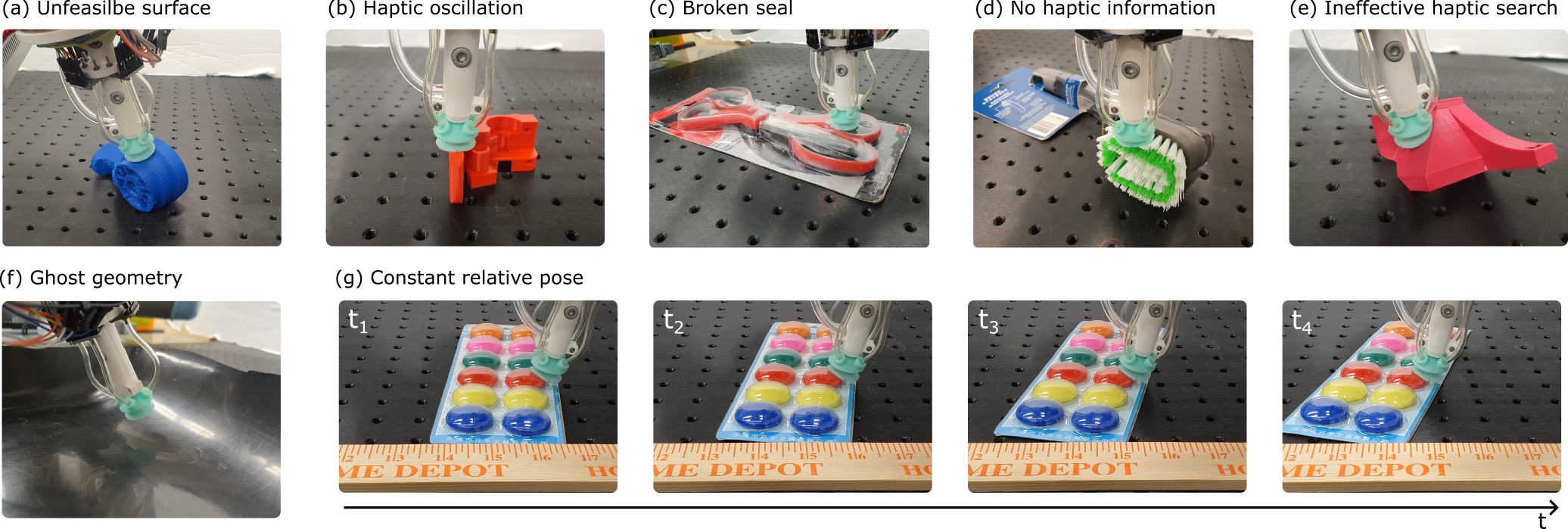}
    \vspace{1pt}
    \caption{Representations of the failure modes for the Smart Suction Cup observed during the bin-picking experiments.} %\jp{I want to get any feedback about this failure taxonomy. Ueven surface, Edge contact, and Broken seal are from AutoLab's paper, and add the case of (d) Ghost effect and (e) Contant relative pose. Please give me any other name of it if you have. For ghost effect, other candidates are reflect/transparent, hallucination, etc. For (e), coupled motion, sliding/tilting, ect.}}
    \label{fig:Failure_taxo}
\end{figure*}

The proposed haptic search methods are labeled $\alpha_1$ to $\alpha_5$ and $\alpha_{1\&5}$. Results show that $\alpha_2$ provides the highest number of successful picks per trial, with an average of 12.6 $\pm$ 4.16. Lateral positioning ($\alpha_1$) and rotational alignment ($\alpha_5$) show reduced results similar to one another, with 8.2 $\pm$ 5.17 and 8.8 $\pm$ 3.70 successful picks, respectively. $\alpha_3$ results in an average of 8.2 $\pm$ 3.83 successful picks and $\alpha_4$ provides successful picks (9 $\pm$ 6.86), but with the largest standard deviation. For the performance of the alternating haptic search method $\alpha_{1\&5}$, it shows the lowest successful picks of 6.8 $\pm$ 1.92 among all the haptic search methods evaluated. 
Overall, these results demonstrate the effectiveness but also the between-trial variability of the proposed haptic search methods. Out of these methods, $\alpha_2$, which predominantly performs lateral search but with some rotational alignment, best improves the success rate of bin picking by the robot system. However, between trial variability indicates that the potential benefits of haptic search is sensitive to initial bin state.

We can then compare the autonomous haptic search methods with the experimental control cases. In the region between 0 and 5 pick attempts, there is little difference between all eight methods. This indicates that success is driven by the GQCNN method, mostly because we attempt the grasping pose with the highest quality value first. The methods diverge in performance after 5 attempts, where the GQCNN and $\alpha_1$ methods show lower performance than the other six methods. At 25 or more bin pick attempts, all six haptically-driven methods outperform the two experimental control methods. This indicates that autonomous haptic search methods are helpful to expand achievable grasp points, to now include those that GQCNN alone is unable to accurately predict.

\revision{Here, we quantify how much fine-tuning is executed through haptic search on average. Among the successful haptic search trials, across all six haptically-driven methods, the mean cartesian displacement from the initial pose was 4.8 mm with %a median of 4.1 mm and 
a maximum of 13.9 mm. 
% \revision{Mean haptic search duration was 4.8 s with a median of 4.1 s and a maximum of 13.9 s.} % revisions did not ask for duration, but good for us to know for reference.
Mean path length was 8.7 mm %(1.9x mean displacement) 
with %a median of 7.3 mm and 
a maximum of 32.7 mm. 
Mean angular displacement was 5.9$^{\circ}$ with %a median of 3.7$^{\circ}$ and 
a maximum of 25.2$^{\circ}$. Mean angular distance traveled was 6.8$^{\circ}$ %(2.0x mean angular displacement) 
with %a median of 4.0$^{\circ}$ and 
a maximum of 39.1$^{\circ}$.}
% \revision{Mean ratio between path length and displacement was 1.9x with a median of 1.9x and a maximum of 3.2x. Mean ratio between angular distance traveled and angular displacement was 2.0x with a median of 1.0x and a maximum of 25.0x.}
% \revision{The haptic search paths are consistently longer than the shortest possible path.}
%\todo{Note all haptic methods better than both control methods, but only at higher pick attempts.}
%\revision{}

\section{Discussion}
\label{sec:discussion}

\subsection{Sensor characteristics}

Through varying the lateral displacement and yaw of the cup against a flat plate edge and varying orientation with domes of different sizes, we characterized the scale and types of pressure signals that the Smart Suction Cup produces. We also demonstrated how these raw signals are interpreted using our proposed haptic search procedure.
However, plates and domes represent primitive shapes.
The complexity of object geometries in real-world scenarios, with a combination of vertical and horizontal flows, will likely impact the haptic search effectiveness of the suction cup, making it challenging to identify suitable direction vectors. This may help us to understand why, at times, we observed certain unproductive haptic behaviors emerge during the bin-picking task. %a reduction in the accuracy of our proposed haptic search method.

 %The elimination of indistinguishable data, where pressure differences between chambers fall below the threshold pressure, results in a slight decrease in direction errors. In \cref{fig:lateral}e ii, thresholding removes most data of estimated yaw angles above the 45$\degree$ boundary line. While some of the data still exceed the boundary, most of it is within a 90$\degree$ boundary, indicating that the estimated direction vector has a component along the true direction vector, thus reducing the lateral offset. Consequently, the haptic search can succeed as long as it starts with a direction error of lower than 90$\degree$.

%\subsection{Rotational Alignment sensor characterization}
We found in sensor characterization tests that thresholding reduced direction error, by eliminating cases where pressure differential measurements are too low to produce reliable estimates when sensor noise starts to dominate.
At the same time, it is unlikely that perfect prediction accuracy is essential in effectively deploying Smart Suction Cup haptic search. Specifically, the prediction accuracy appears to improve as the cup gets closer to a successful grasp. %Considering the signal over time when coupled with searching motions 
During haptic search, if, as a result of noisy signals due to low pressures, % or other sources of error, 
the cup randomly reaches any state where a more accurate prediction can be better made, then the behavior will converge on a successful grasp over time. We posit that this will be especially true if, on average, predictions start from a place that are within 90$^{\circ}$ of the true direction vector. %We note here that having a few time points will still likely result in a successful grasp, since on average, the direction of motion will be nominally correct. 
%Therefore, we should also consider the mean and median direction error at each offset. Doing so yields higher misalignment tolerance with up to 35$\degree$, 45$\degree$, 37$\degree$ for domes with radii of 15mm, 20mm, and 40mm, respectively. 
In future work, conducting closed-loop control experiments, rather than stationary sensor characterization, would identify the highest possible offsets for which haptic search still yields a successful grasp, including on a wider variety of object shapes.

In the lateral search case and especially the %Characterization results from 
rotational alignment case, we find that the compliant material and the bellows of the suction cup allows it to engage with objects even with postural errors to some extent. However, for objects with high curvature and critical features such as holes, the inherent tolerance of the suction cup may not be sufficient. In such cases, our proposed haptic search method is expected to enhance the operational tolerance even when the vision system fails to capture those features accurately.

\subsection{Bin-picking observations}

Bin-picking experiments suggest that a physical search phase after contact is made %haptic search 
can improve grasp success, especially when employing autonomous haptic search methods that respond to measured contact conditions. %when compared to random searching. 
The fact that all haptic search methods tested provided some increase in picking success rate as compared with experimental controls, including with random searching, shows how responding to contact pressures, even with a simple model-based controller, holds potential thus motivating ongoing investment in the Smart Suction Cup capability. 
We used a single suction cup prototype throughout all of these bin-picking-experiments, representing at least 1316 autonomous grasp attempts, without incurring damage to the cup or needing replacement. The Smart Suction Cup design, where the cup is fabricated in a single-step casting process and electronics are remote from the cup, thus appears to provides reliable and physically robust performance. 

We saw the biggest performance increase with the $\alpha_2$ haptic search method, whose motion is a mix of lateral positioning with a bit of rotational alignment. Though it matches our expectations that a coupled motion would yield better results than purely sliding ($\alpha_1$) or rotating ($\alpha_5$), because most objects have both edges and curves, it is less obvious why $\alpha_2$ outperforms $\alpha_3$ and $\alpha_4$. 
A possible theory is that the rotational alignment search counteracts the lateral search, so finding the optimal tuning between them is required. When the suction cup has partial contact, the lateral search attempts to reinforce contact on the contacted side by moving towards it, while the rotational alignment loosens the contact side and attempts to make balanced contact over all channels. Therefore, an appropriate balance between the two modes should be adjusted. We believe that $\alpha_2$ provides the best balance among the five presets in general, but each geometry may require a different optimal balance between the two modes. We leave this local, object-specific controller optimization as a future work. 
%One possible theory is that the grasp poses from GQCNN are suitable for handling curved surfaces but may not be ideal for geometries that require fine adjustments in the lateral direction with the vision system we used.

During the bin-picking trials with autonomous haptic search, we observed different common grasp failure modes. We classify them into seven categories, as shown in \cref{fig:Failure_taxo}: %. The Smart Suction Cup with proposed haptic searches is unable to achieve suction grasps due to:
\begin{enumerate}[(a)]
    \item \textit{Unfeasible surface:} Haptic search starts at an infeasible surface, where possible grasp poses are beyond the searching boundary.
    \item \textit{Haptic oscillation:} Haptic search oscillates in a region where haptic information makes the cup move back and forth without converging to a graspable point.
    \item \textit{Broken seal:} The contact wrench applied to the cup is too large to lift an object. This typically occurs when the suction cup tries to grasp a heavy object from the edges, also reported in \cite{sanders2020non}.
    \item \textit{No haptic information:} The suction cup cannot get any distinguishable haptic data from a surface, such as the bristles of the brush (P$<$10 Pa).
    \item \textit{Ineffective haptic search:} A surface is feasible and haptically searchable, but the system uses an ineffective behavior. The example shows a case where the suction cup is using lateral positioning but would benefit more from rotational alignment.
    \item \textit{Ghost geometry:} Reflective and/or transparent materials yield artifacts, resulting in ghost geometries in a depth image. The example in the figure shows the suction cup is trying to grasp in the air because the light from the ceiling is reflected on the bin surface.
    \item \textit{Constant relative pose:} During haptic search, a loose object can be pushed such that the relative position between the cup and object remains unchanged despite robot motion. Given the new position of the object, %If an attempt to grasp an object fails, it is pushed and a new attempt is made. The 
    the next attempt may consider the same grasp point as a valid candidate %the previously failed grasp point of the object as a new candidate, 
    as its pose in the world frame changed.
    %Objects are not fixed in position and orientation. Consequently, the relative pose between the cup and an object doesn't change enough to succeed in a haptic search.
\end{enumerate}

%\hs{I would add images of specific circumstances or objects here to elucidate specific conclusions with examples.}

Several of these error types occur because the vision-based grasp planner initializes the grasp at a point in which a suction grasp is locally impossible. The cases in \cref{fig:Failure_taxo} (a), (c), and (f) are not recoverable using contact condition condition sensing. % alone with the haptic search methods. 
To combat these, the camera and/or visual planner performance would need to be improved. However, for cases in \cref{fig:Failure_taxo} (b), (d), (e), and (g), new adaptive haptic search controllers designed to identify and overcome such failure cases could further improve grasping in future work.
\revision{For instance, the failure mode (g) may be effectively addressed through a jumping haptic search approach. In this approach, the suction cup retracts from an object and then re-approaches with an adjusted pose. This prevents the suction cup from exerting continuous pressure on an object while making pose adjustments}.
We recommend also coupling vision with the haptic search process. For example, camera information could be used to select appropriate haptic search methods in response to case (e). Or vision could identify object movement in (g) to adapt behavior on the fly; %In the present work
for example, in \cite{huh2021}, we propose that one could dynamically reduce the suction pressure of the vacuum in order to achieve more gentle sliding over objects.

% \hs{From observations of these failure modes...}
% \jp{
% Note that a failure case can be a combination of the above taxonomy (specific examples?)
% Some failure modes can be overcome with an improvement of vision.
% Constant/diminished relative pose is the most critical failure mode for our proposed haptic search because the key concept of haptic search is to adjust pose.
% An ineffective haptic search can be improved in the future iteration by using adaptive weight value, or smart weight value selection.
 % Our preliminary test with fixed objects found that $\alpha_1$ had the highest number of successful haptic searches (data are not shown here). However, during the bin-picking experiment, the object continuously changes during the haptic search, making a small amount of rotational alignment the most helpful.
% }

In the present work, we made the deliberate choice not to re-train the GQCNN algorithm for our particular robot system, resulting in overall low planner performance. In our case, we use a different camera, robot arm, object set, room/lighting, gripper, and bin from the ones used in training. The purpose of this choice is to generate a scenario that emulates a quick-adopt case for such technology, since generalizability is an ongoing challenge for such planning algorithms \cite{dasari2019robonet}. %discuss how this is a deliberate decision with implications for real-world adoption and generalizability. 
The present work therefore shows that the use of a Smart Suction Cup can be one tool in ameliorating errors that arise % bridging adoption, and improving overall system performance 
specifically in previously unseen systems.
%However, we also recognize the low performance of our baseline GQCNN method, as compared with prior implementations. We suspect that the role of haptic exploration is most useful when such planners under-perform, highlighted in the present study.
Future work will investigate how planner optimization and hardware selection (e.g., higher spatial resolution camera) affects the role of autonomous haptic search.

\begin{figure}[b!]
    \centering
    \includegraphics[width=0.8\linewidth]{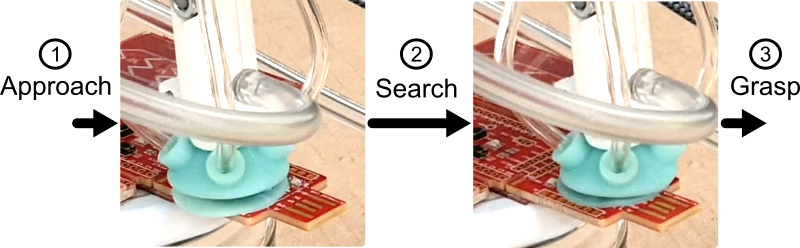}
    \vspace{2pt}
    \caption{An example of lateral haptic search, $\alpha_1$, on a stationary PCB adhered to a surface. \textcircled{1} ``GQCNN'' guides to the pose where there are several via holes. \textcircled{2} The suction cup adjusts its lateral pose given its pressure readings. \textcircled{3} The haptic search succeeds in grasping the PCB within 2 seconds.}% \todo{reduce this to 3 images only, can zoom in more to the suction cup if needed.}}
    \label{fig:PCB}
\end{figure}

\subsection{Printed Circuit Board demonstration}

%Brainstorming:
In the bin-picking experiment, the tested objects all had at least one smooth graspable surface for the suction cup to grip. %This is a standard way to compare results with existing testing procedures used across relevant studies. % and is arguably a more realistic task than trying to grip objects that are fixed in place. %Because the objects can move, this results in the failure case in \cref{fig:Failure_taxo}g. %, where the object slides to negate the potential benefit of the haptic search procedure. %In \cite{huh2021}, we propose that one can reduce the suction pressure of the vacuum in order to ease sliding over non-stationary objects without pushing them. %Additional behaviors may also be used to combat such issued such as briefly lifting the gripper and setting it down again in a new position, rather than sliding. Addressing this failure case is left to future work. 
%This object sliding issue goes away on stationary, fixtured objects. 
However, some real-world objects have bumpy surfaces without any obvious continuously-smooth regions. For example, a Printed Circuit Board (PCB) with Integrated Circuits (IC) soldered on it and via holes might prevent the use of a suction cup, if the cup would fail to grasp at most surface locations. %This is especially challenging if there is no \textit{a priori} model for a given PCB, e.g., in a recycling task. 
However, haptic search behaviors can still enable the grasping of such surfaces, adapting around local features to achieve a seal. %uniquely emerge on immovable objects.
%Because of this, additional behaviors emerge from the implemented haptics search procedure such as edge following. 
To demonstrate this behavior, we fix a printed PCB to the table and allow the cup to search for a grasp point using only lateral positioning, or $\alpha_1$. 
%This is a particularly challenging object for a suction cup as vias and integrated circuits across the surface can break the seal. 
\cref{fig:PCB} shows how the cup is able to find a successful grasp point over one of the IC's. %In fact, when we perform testing on a single object (such as the elf) we find that haptic exploration can at times greatly improve successful gripping...
%\hs{Can we include the PCB demo or any demos on fixed objects?}
Future work will measure to what extent the cup can respond productively on surfaces with different types of porosity and rugosity profiles for real-world applications. %how object fixturing can improve haptic searching methods in appropriate applications.

\revision{In {\cref{fig:lateral}}, the directional errors in lateral search on a flat, smooth plate commonly reach almost 20$^\circ$ for the best case lateral offsets between 8 and 13 mm. These errors may appear unsatisfying and at times result in longer searching paths than desired. This directional error provides one reasonable explanation for the edge-following behavior that emerges at Grasp point 3 in Supplementary Video. However, the PCB demonstrations show how this error does not necessarily result in overall failure during smart suction haptic search; the controller continues to adjust its directional estimate every 0.5 mm as it moves, ultimately leading to a successful grasp. Regardless, future work should investigate how performance -- such as time and distance to successful grasp -- may be optimized through cup and algorithm design.}

%\hs{I would add images of specific fixed-object circumstances to elucidate specific conclusions with examples.}

%\subsection{Limitations and future work}

\section{Conclusion}
\label{sec:conclusion}

The four-chamber cup design of the Smart Suction Cup, with remote pressure transducers, provides a reliable solution for generating differential airflows and protecting sensitive electronics from physical damage. In this work, our proposed autonomous haptic search method -- a model-based approach for estimating lateral positioning and rotational alignment -- enables the suction cup to adjust to a successful pose for suction grasping, effectively increasing tolerance to positioning or misalignment error induced by errors from a vision-based grasp planner. %Our experiments demonstrated that the Smart Suction Cup with haptic search algorithms outperformed a previous vision-only suction grasp method in bin-picking tasks. 
%The potential applicability of t
The Smart Suction Cup holds the potential to improve gripping in various scenarios that already deploy vacuum grippers, such as recycling facilities, warehouses, manufacturing, and logistics robots. %It also presents a promising solution for improving the universality of such gripping technology and translating them to new applications which have previously been too varied or adversarial for prior adoption. %Therefore, our proposed device and associated haptic search methods have the potential to contribute significantly to the development of a robotic system.

\subsection{Future work}

%\hs{For very high-level future directions}

This study presented a single implementation of the Smart Suction Cup and one particular model-based approach to generating haptic searching behaviors in response to pressure readings.
In future work, we seek to explore new soft cup designs to both improve gripping performance while studying how parameters, like the number of chambers, affect sensing.
%Design improvement. gripping performance is affected by altering the number of chambers will be a part of future work.
% Next steps should also include methods of optimal or learned control for mixing of lateral positioning and rotational alignment.
Next steps include optimization and learning-based approaches for \revision{sensor characterization and} mixing lateral positioning and rotational alignment.
These adaptive methods may be informed by visual and haptic information, for example. %A smart selection of weight value
Finally, the ultimate goals of this line of work is to explore the adoptability and lifetime of such technology in real-world application.

% Astrictive grasp wrench limits are influenced by porosity, texture and other non-idealities of the surface and cannot necessarily be characterized by a wrist load cell and visual sensors alone.}

%Future work 
% - estimate the maximum pull force from lowered vacuum exploration mode.
% - adaptive control to prevent detach failure.
% W

\section{Acknowledgments}
%T.M.Huh  

This work is supported by 
InnoHK of the Government of the Hong Kong Special Administrative Region via the Hong Kong Centre for Logistics Robotics
%the InnoHK of the Government of Hong Kong via the Hong Kong Centre for Logistics Robotics 
and by the University of California at Berkeley. The authors thank the members of the Embodied Dexterity Group.
% M. Danielczuk was supported by the National Science Foundation Graduate Research Fellowship Program under Grant No. 1752814. The work of M. Li was
% supported by the NASA Space Technology Research Fellowship, under Grant
% \#80NSSC19K1166.
%© 2023 IEEE.  Personal use of this material is permitted.  Permission from IEEE must be obtained for all other uses, in any current or future media, including reprinting/republishing this material for advertising or promotional purposes, creating new collective works, for resale or redistribution to servers or lists, or reuse of any copyrighted component of this work in other works.

%\hfill \break
%\clearpage

% \input{5_HapticExploration}
% \input{6_DynamicDetachMonitoring}
% \input{6_AdaptiveGrasping.tex}
% \input{7_Conclusion}
% \input{2_Theory}
% \input{3_Design}

\bibliographystyle{IEEEtran}

\balance  % We need this line for the correct ordering of reference. ( this line is in "balance" Package

\bibliography{IEEEabrv,Biblio}   % Reference are in "Biblio.bib" file.
%\vspace{\baselineskip}

%\input{100_Bio.tex}

\end{document}